\documentclass[journal]{IEEEtran}
\usepackage[percent]{overpic}
\usepackage{color}
\usepackage{dsfont, caption}
\usepackage{amsmath, graphicx}
\usepackage{amsfonts, epsfig, mlspconf, algorithm2e, subfigure, color}
\usepackage{siunitx}
\usepackage{url}

\newcommand{\revised}[1]{#1}
\newcommand{\specialcell}[2][c]{%
  \begin{tabular}[#1]{@{}c@{}}#2\end{tabular}}
\newcommand{\colw}{0.3}
\newcommand{\figw}{0.45}

\begin{document}
\title{Deep Learning for Surface Material Classification Using Haptic and Visual Information}
\name{ Haitian Zheng$ ^1 $, Lu Fang$ ^{1,2} $, Mengqi Ji$ ^2 $, Matti Strese$ ^3 $, Yigitcan \"Ozer$ ^3 $, Eckehard Steinbach$ ^3 $}
\address{
	$ ^1 $ University of Science and Technology of China, \small{\textit{zhenght@mail.ustc.edu.cn}}\\
	$ ^2 $ Hong Kong University of Science and Technology, \small{\textit{\{eefang, mji\}@ust.hk}}\\
	$ ^3 $ Technische Universit\"at M\"unchen, \small{\textit{\{matti.strese, yiit.oezer, eckehard.steinbach\}@tum.de}}}

%


\maketitle
\begin{abstract}

When a user scratches a hand-held rigid tool across an object surface, an acceleration signal can be captured, which carries relevant information about the surface material properties. More importantly, such haptic acceleration signals can be used together with surface images to jointly recognize the surface material. In this paper, we present a novel deep learning method dealing with the surface material classification problem based on a Fully Convolutional Network (FCN), which takes the aforementioned acceleration signal and a corresponding image of the surface texture as inputs. Compared to the existing surface material classification solutions which rely on a careful design of hand-crafted features, our method automatically extracts discriminative features utilizing advanced deep learning methodologies. Experiments performed on the TUM surface material database demonstrate that our method achieves state-of-the-art classification accuracy robustly and efficiently.
\end{abstract}


%
\IEEEpeerreviewmaketitle

\section{Introduction}
\label{sec:intro}

With today's sensor technology, a wide variety of data types can be captured. Understanding the sensory data and recognizing objects is becoming an important research topic. Recent work in object recognition \cite{TMM-dl-fusion} and indoor scene recognition \cite{TMM-dl-fusion2}, for instance, demonstrate that combining visual information with depth sensory input improves the classification performance. Taking material surface classification as another example, besides camera sensors, the easily-accessible acceleration sensor is able to record vibration signals when it slides over a surface. Such vibration signals capture information about the material properties of the surface, and also reveal rich haptic attributes. These haptic signals are complementary to visual input, providing an opportunity towards a better material classification scheme. In this paper, we investigate such multimodel material surface classification problem which involves both visual and haptic inputs.

Surface material classification has recently gained increasing interest. When a rigid tool slides on a surface, the resulting vibrations of the tool contains information about characteristic properties of the surface \cite{Haptic:TUMdata}, such as hardness and roughness of the material. There has been increasing interest to recognize surface materials using robots \cite{Haptic:TUMdata}\cite{Haptic:Mengqi}\cite{Haptic:Recognition2}\cite{Haptic:Recognition3} and to recreate the haptic feel of real surfaces. However, surface material classification from tool / surface interaction data becomes particularly challenging if free-hand movements are considered.


\begin{figure}[htbp]
	\centering
	\includegraphics[width=8.5cm]{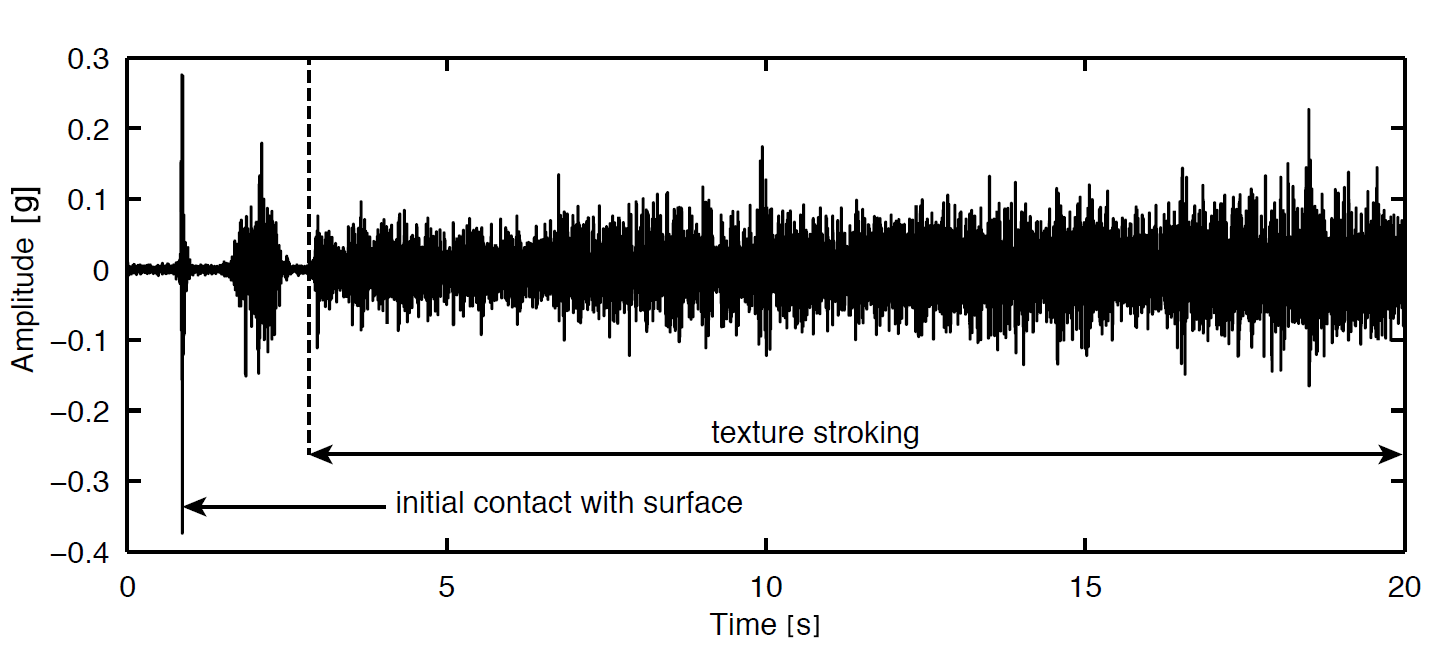}
	\caption[Haptic Datatrace]{Surface texture data trace, reproduced from~\cite{Haptic:TUMdata}. It can be observed that higher velocities (left to right) increase the signal power and variance, while the exerted normal force is held constantly.
		\label{fig:rawsignal}}
\end{figure}


\revised{When a human strokes a rigid tool over an object surface, the exerted normal force, the tangential scan velocity, and the angle between the tool and the surface might vary during the surface exploration and between subsequent exploration sessions. These scan-time parameters strongly influence the nature of the recorded acceleration signals~\cite{TextureClassification:Romano2012}. Fig.~\ref{fig:rawsignal} shows an example of an acceleration data trace, where the scan velocity is linearly increasing and reveals, how this change influences the data trace with regard to its signal power and variance. The variability of the acceleration signals thus complicates the texture classification process.}

Before surface classification using acceleration data (captured while interacting with the surface) emerged \cite{Haptic:TUMdata}\cite{Haptic:Mengqi}\cite{Haptic:Recognition1}\cite{Haptic:Recognition2}\cite{Haptic:Recognition3}, \revised{a significant number of previous works have focused on using photos of material surfaces to classify the material types.} These approaches mainly rely on hand-crafted image features including locally binary pattern (LBP) features \cite{TextureImage:binaryPatterns}, filter bank features \cite{TextureImage:randomFeatures}\cite{TextureImage:filterbank}, co-occurrence matrix based features \cite{TextureImage:cooccurrence} etc, in combination with appropriate machine learning tools to distinguish the different texture types.

Designing specific features requires specific domain knowledge. Recently, Convolutional Neural Networks (CNN) have become a popular tool for pattern recognition, allowing better features being extracted automatically. In the context of surface classification, \cite{TextureImage:CNN}\cite{TextureImage:CNNThesis} aim to classify texture image patches by training CNNs, \revised{\cite{TextureImage:Deep filter banks} designs fisher-vector descriptors of texture images using the ImageNet pretrained CNN, \cite{TextureImage:Using filter banks in CNN} uses CNN to learn texture image filter banks and \cite{TextureImage:Dynamic texture} designs dynamic texture descriptors utilizing the ImageNet pretrained CNN}. Meanwhile in the context of haptic classification, our previous work \cite{Haptic:Mengqi} proposes an auto-encoder pre-trained CNN for classifying texture haptic segments\footnote{\cite{Haptic:Mengqi}, a preliminary version of this work, has appeared at MLSP 2015. It mainly focuses on directly dealing with the one-dimensional haptic signal without any preprocessing procedures. As an extension, this work dives into a more efficient solution with hybrid-inputs (haptic as well as image signals).}. However,

\begin{itemize}
  \item compared to the texture signal (image or haptic) which could be of arbitrary size, CNN has a relative small receptive field. In order to reconcile the disagreement of the two mentioned sizes, an inefficient sliding window based approach needs to be adapted, which might jeopardize the efficiency of a real-time application.
  \item regardless of the significant progress in haptic (acceleration)-only or image-only classification, there are rarely approaches dealing with hybrid data input. For instance, two types of surface materials with similar image appearance can lead to completely different acceleration data, and vice versa. In such cases, better classification performance can be achieved by utilizing both haptic (acceleration) data and image data. 
\end{itemize}

To overcome the inefficiency of CNN + sliding windows scheme, a Fully-Convolutional Neural Network (FCN) approach is adapted in our work. FCN \cite{FCN} is a special type of convolutional neural network which replaces fully connected layers with convolutional layers with a $1 \times 1$ convolution kernel. Without fully connected layers, the FCN is able to take input of arbitrary size, and outputs label predictions at every receptive field. More importantly, compared to the approach of `CNN + sliding window', the FCN can be trained and tested more efficiently.

Different from previous approaches in adapting FCN for vision tasks, we propose a systematic FCN scheme for recognizing surface materials from hybrid data (haptic signals and images). The FCN for haptic data recognition is trained using concepts developed for speech recognition, as haptic data share similar characteristics with speech data \cite{Haptic:TUMdata}, \cite{Haptic:Recognition1}; the FCN for image-based texture recognition is trained by fine-tuning the network weights from \cite{CNN:Alex}, inspired by transfer learning \cite{CNN:transfer}. Afterwards, additional hybrid networks further integrate the haptic/visual features for better classification performances. Experiments conducted on publicly available surface material datasets \cite{Haptic:TUMdata}\cite{TextureImage:Kylberg}\cite{TextureImage:KTH} demonstrate the superior performance of our scheme in terms of both efficiency and accuracy for surface material classification. \revised{While the concurrent work \cite{haptic:PennSimilar} also applies deep learning for haptic/visual hybrid input-based surface classification, we would like to point out that
1) We are eventually handling very different dataset from \cite{haptic:PennSimilar}. In \cite{haptic:PennSimilar}, the material object is supposed to be at the center of a fixed-size image, and the haptic trace records the fixed-length signal when robot gripper explores specific procedures such as squeeze, hold, etc. While the TUM image/haptic-texture dataset contain repetitive patterns, and can be of arbitrary size. 2) Given the fix-size dataset, \cite{haptic:PennSimilar} developed a one-shot CNN classification framework. To deal with the dataset with repetitive patterns and arbitrary size, we develop a `FCN + max-voting' framework, where FCN improves the naive `CNN + sliding windows' approaches by speed, and the max-voting further boosts the FCN prediction accuracy. 3) Although \cite{haptic:PennSimilar} tested image/haptic fusion, they did not perform the joint training of fusion network. In this paper, with the help of the larger dataset, the joint training is possible and performed.}

The remaining paper is structured as follows. In Section \ref{sec:Related}, we discuss the related work and present CNN and FCN. In Section \ref{sec:method}, the details of our method are elaborated. Section \ref{sec:experiment} describes the surface classification experiment on the TUM dataset and discusses the additional results. Finally, Section \ref{sec:conclude} concludes the paper.

\section{Related Work}
\label{sec:Related}


Convolutional neural networks (CNN), a type of trainable multistage feed-forward artificial neural network that extracts a hierarchical feature representation, are a powerful tool for both image and speech recognition. A typical CNN consists of convolution layers, pooling layers, and fully connected layers. The main ingredients of a CNN can briefly be summarized as follows:

\begin{itemize}

\item{Convolution layers} extract feature maps from the input by applying consecutive convolution operations between the input and trainable kernels, followed by a non-linear activation function.

\item{Pooling layers} usually follow convolution layers, aiming to reduce both the dimensionality and translation sensitivity of the input feature maps. For image recognition, pooling layers significantly boost spatial translation invariance, while for spectrogram recognition pooling layers lead to temporal-frequency invariance.

\item{Fully-connected layers} are usually used at the ending stage of a CNN, providing more flexible feature mapping. In a fully-connected layer, the input feature vector is linearly converted into a new feature vector before being fed into a non-linear activation function.

\item{Softmax layer} is usually used for handling the multiple label regression problem. At the end of the neural network, the softmax layer outputs the normalized exponential of the input vector, which indicates the probability of each label.

\end{itemize}

The activation function of the CNN can be chosen among sigmoid function, tanh function, and rectified linear (ReLu) function \cite{CNN:reLu}, among which, ReLu becomes more and more popular due to its efficiency for training and effectiveness for improving the classification performance. Additionally, dropout regularization \cite{CNN:dropout} is commonly used after fully-connected layers, which significantly reduces co-adaptation between features, and hence prevents over-fitting and boosts the classification performance significantly.


CNNs have experienced great success in a wide range of vision tasks, including image recognition~\cite{CNN:Alex}, \cite{CNN:VGG}, detection \cite{CNN:Detection1}, \cite{CNN:Detection2}, segmentation \cite{FCN} and many specific applications such as image aesthetics assessment \cite{CNN:Aesthetics} and eye fixations prediction \cite{CNN:Eyefix}. In image recognition specifically, with the help of large datasets (i.e., ImageNet), CNN methods \cite{CNN:Alex} \cite{CNN:VGG} \cite{CNN:GoogLeNet} have taken over the lead in large scale visual recognition challenges (ILSVRC) since 2012. For a small dataset, learning the millions of parameters of a CNN is usually impractical and may lead to over-fitting. CNNs, however, still successfully show their power -- studies on transfer learning \cite{CNN:transfer} \cite{CNN:transfer-DECAF} \cite{CNN:transfer-Visualizing} show that the trained CNN model for one specific vision task usually learns a good representation of natural images, which works for other visual recognition tasks as well. Inspired by the success of transfer learning, we investigate in this paper how to handle the surface texture recognition task using the relatively small TUM image dataset \cite{Haptic:TUMdata}.

CNN has also been shown to be a powerful tool for speech recognition. Recent works \cite{Speech:CNN1} \cite{Speech:CNN2} \cite{Speech:CNN3} \cite{Speech:CNN4} \cite{Speech:CNN5} show that CNNs notably outperform fully-connected deep neural network (DNN). The superior performance is attributed to the property of temporal-frequency translation invariance inherent with CNN \cite{Speech:CNN3}. In addition to speech recognition, CNNs are applied to acoustic recognition tasks such as music genre classification \cite{Acoustic:music}, music onset detection \cite{Acoustic:music-onset} and music adversaries \cite{Acoustic:music2}. Motivated by the previous work on 1-D speech signal recognition, and evidence in \cite{Haptic:TUMdata}\cite{Haptic:Recognition1} which show that the acceleration signals captured during the interaction with an object surface share certain characteristics with speech signals, we investigate a CNN method to deal with our haptic acceleration signal based recognition task.

Though CNNs provide powerful recognition abilities, additional designing is required to handle the special properties of the acceleration signals which describe the surface material. Specifically, the input can be of arbitrary size and contains frequently repeated local patterns. Thus it is more desirable to take local signal segments as the input, rather than the entire signal. Intuitively, the most simple texture prediction pipeline could be: 1) convert the acceleration input into segments using sliding windows; 2) predict every segment with a trained CNN; 3) perform max-voting among multiple CNN predictions. Our experiments show that with a carefully trained CNN, the aforementioned work-flow achieves high prediction accuracy. However, a sliding window based approach is inappropriate for real-world applications, as dense prediction of the CNN is computationally expensive.

In our work, to alleviate this inefficiency, the naive sliding windows approach (step 1 and 2) is replaced by the FCN. FCN is a special type of convolutional neural network which replaces fully connected layers which output $n$ features with convolutional layers with $n$ $1 \times 1$ convolution kernels. The key observation of FCN is that fully-connected layers in a CNN are special convolutional layers: convolutional layers which take a $1 \times 1$ feature map and output another $1 \times 1$ feature map. In contrast to a fully-connected layer which takes fixed-length input merely, the corresponding convolutional layer is generalized to take input of arbitrary size, which is extremely beneficial for extracting dense features at every spatial location.  

Taking this key observation into account, the FCN can replace the fully-connected layers of a CNN by convolutional layers with the same weights. Given the input data of arbitrary size, FCN first performs convolution and max-pooling operation alternatively as a usual CNN, then performs multiple $1 \times 1$ convolution operations (and drop-out), finally provides label predictions at every receptive field. Compared to the sliding windows approach which heavily re-computes feature maps within overlapping ``windows'', FCN is significantly less computationally expensive. As further noted in \cite{FCN}, both training and inference of FCN can be performed {by standard neural network approaches, leading to an efficient and systematic scheme}.


\section{Proposed CNN Scheme with Hybrid Input}
\label{sec:method}
In this section, the proposed CNN scheme with hybrid input will be elaborated by starting with the explanation of hybrid data recording (Section III-A), followed by the surface texture classification schemes (Section III-B) via {\emph{HapticNet}}, {\emph{VisualNet}} and \emph{FusionNet}.

\subsection{Hybrid Data Recording}
\label{subsec:method:haptic}

\begin{figure}[htbp]
	\centering
	\includegraphics[width=7.5cm]{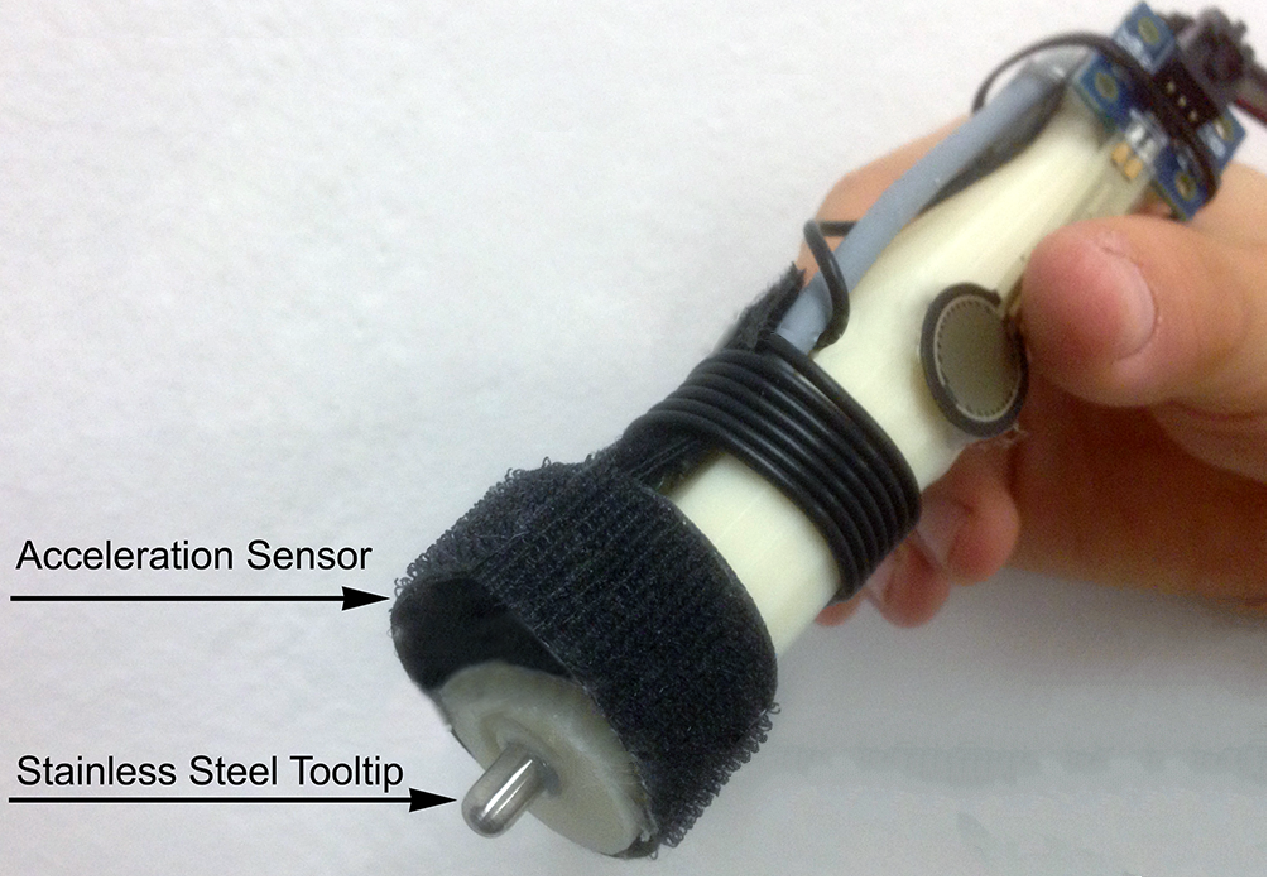}
	\caption[Haptic Stylus]{Haptic stylus used for texture analysis.
		\label{fig:newsensor}}
\end{figure}



\noindent\textbf{Haptic Acceleration Data} In our work, we use the haptic stylus from \cite{Haptic:Recognition1}, which is a free-to-wield object with a stainless steel tool-tip, shown in Fig.~\ref{fig:newsensor}.
In~\cite{Haptic:Recognition1}, a three-axis LIS344ALH accelerometer (ST Electronics) with a range of $\pm \SI{6}{g}$ was applied to collect the raw acceleration data traces. All three axis were combined to one using DFT321 (see~\cite{Haptic:DFT321}). This approach, which preserves the spectral characteristics of the three axes, was adapted in order to have less computational effort in terms of feature calculation.

\vspace{0.3cm}
\noindent\textbf{Image Data} Different from acceleration modality, images are taken in a non-dynamic way of recording. However, differences in distance, rotation, light condition as well as focus also complicate the collection of uniform images for the surface classification task.

\begin{figure}[htbp]
	\centering
		\includegraphics[width=9cm]{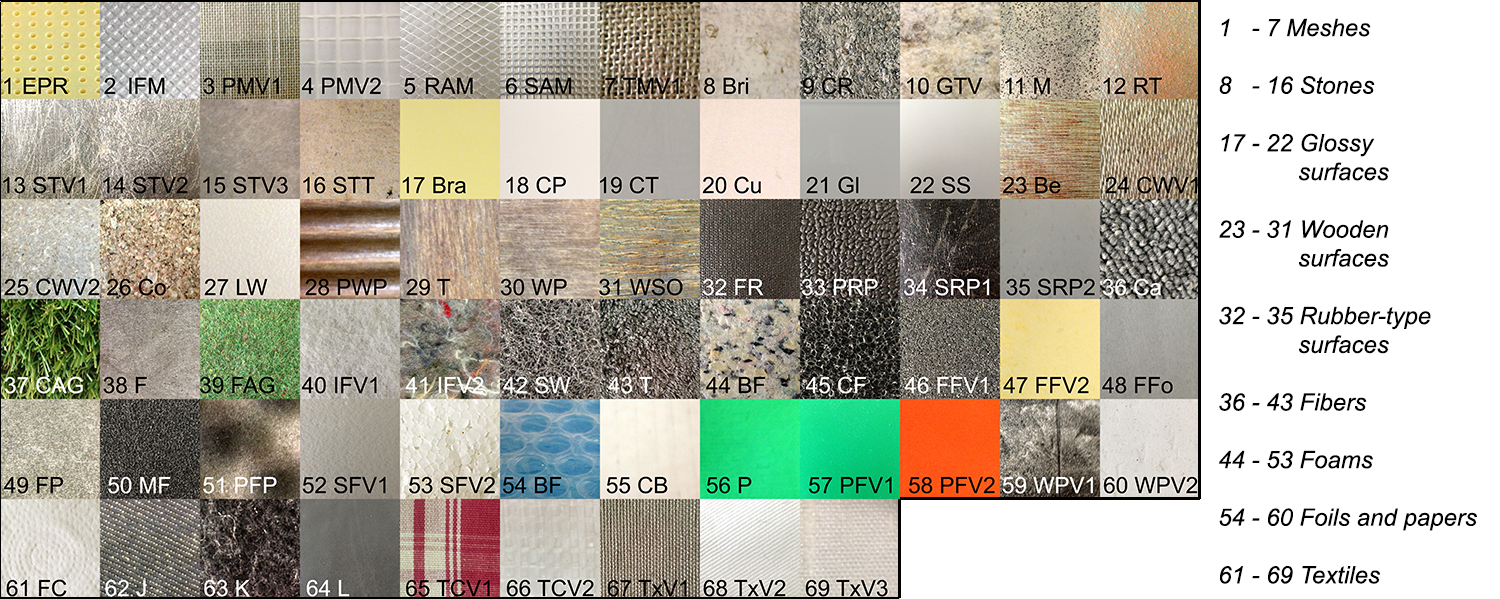}
	\caption[textures]{Materials included in the TUM haptic texture database, freely accessible at \textit{http://www.lmt.ei.tum.de/texture/}
	\label{fig:textures}}
\end{figure}
\begin{figure}[htbp]
   \centering
      \includegraphics[width=8cm]{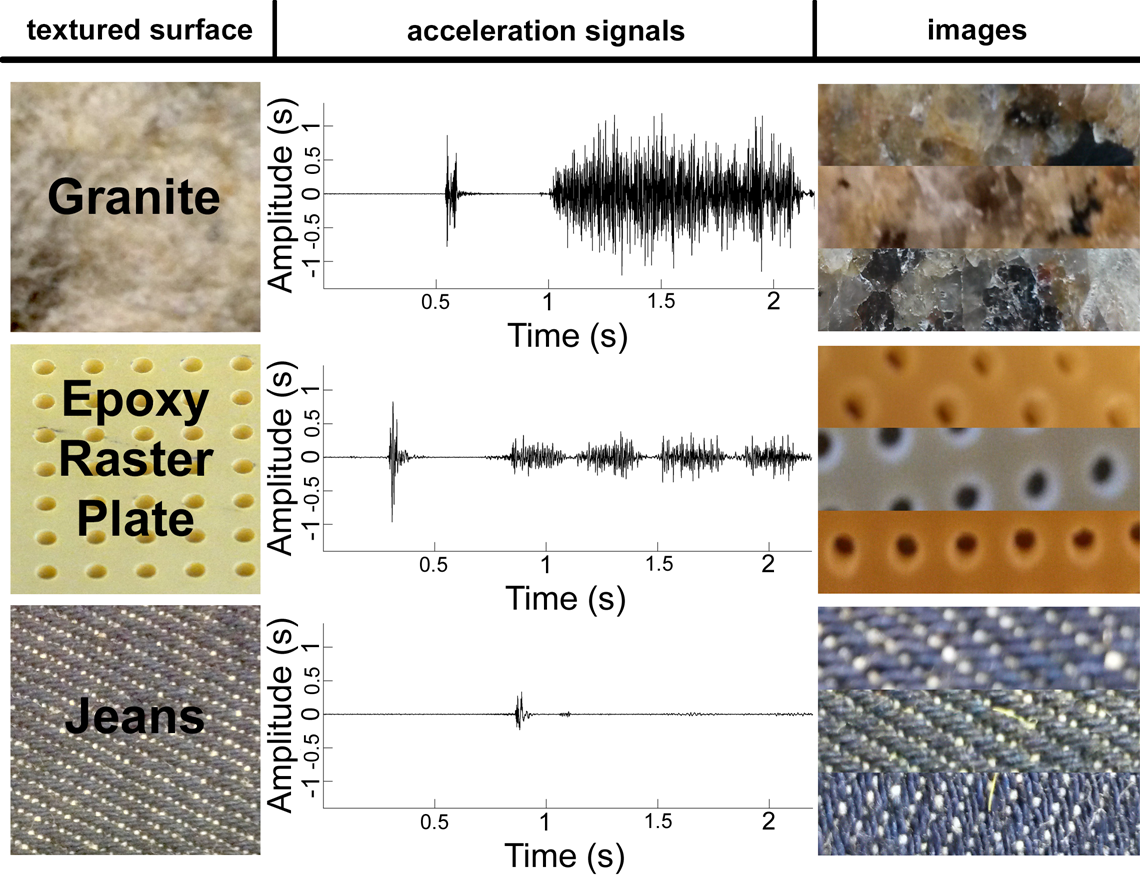}
    \caption[]{Example signal traces of the used image and acceleration data recordings. Arbitrary variations during the recording of the acceleration data traces have been applied. Also, the ten images per textured surface were captured under varying distances, light conditions, focus conditions as well as different camera inclinations towards the surface.
    \label{fig:allmodalities}}
\end{figure}

Fig.~\ref{fig:textures} shows all the used surfaces and their abbreviated names, where the images of the surfaces are taken by the rear camera of a common smartphone (Samsung S4 Mini) and have a resolution of 8 Mega-Pixels. The illumination and viewing conditions differ within the same class of surfaces. For each surface, there are 5 images under daylight and 5 images under ambient light conditions. Viewing direction and camera distance are chosen arbitrarily for each picture, resulting in variations within each individual class. As an example, we choose three textured surfaces and plot the raw acceleration and image data in Fig. \ref{fig:allmodalities}.


\subsection{Surface Material Classification}
\label{subsection:surface_classification}
\begin{figure*}[htbp]
	\centering
	\subfigure[]{\includegraphics[width = 5.0in]{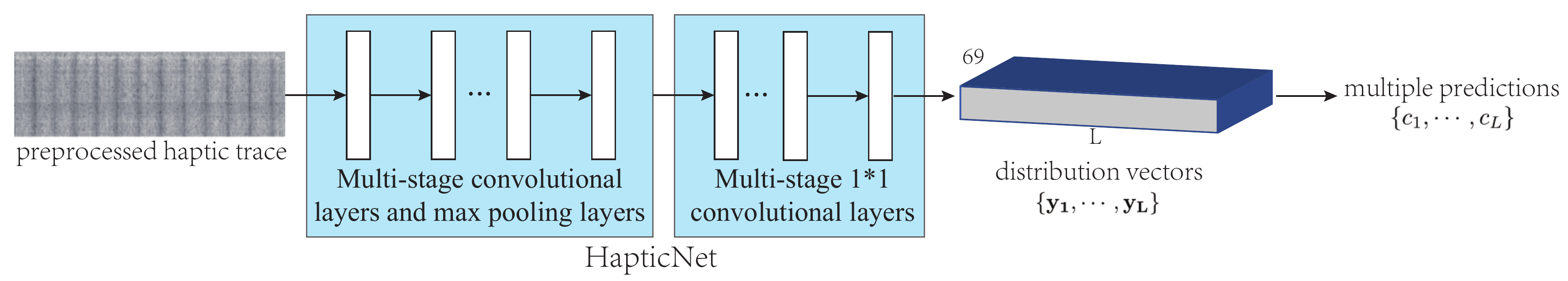}
	}
	\subfigure[]{\includegraphics[width = 5.0in]{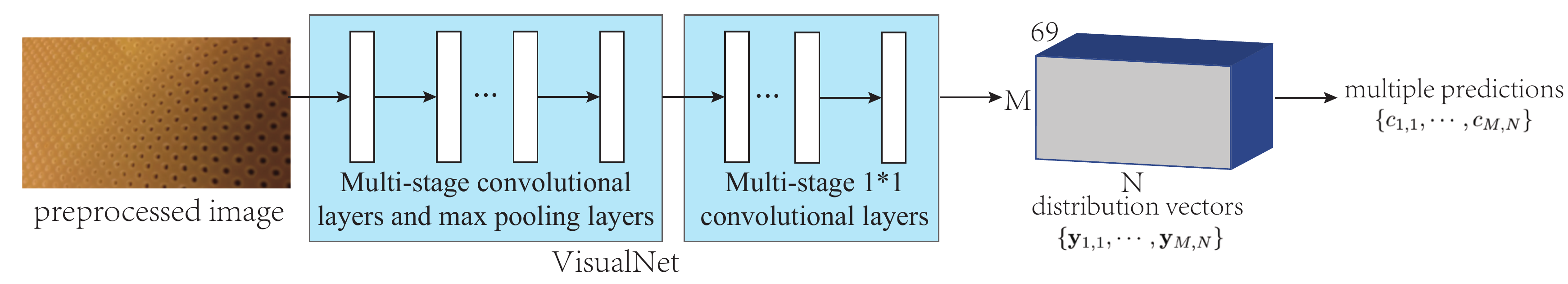}}
\subfigure[]{\includegraphics[width = 7.0in]{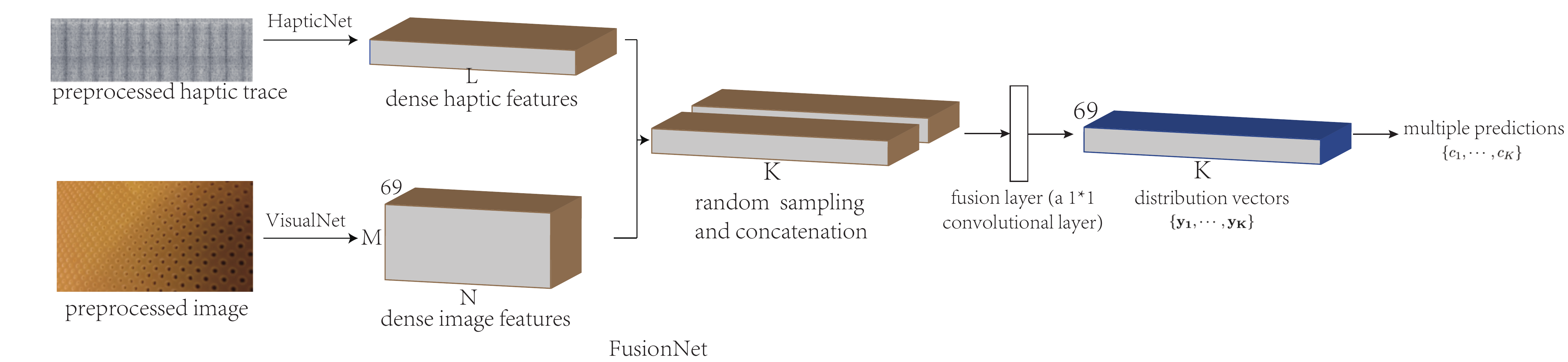}}
	\caption{\revised{The pipeline of the classification scheme. (a) classification of haptic input ({\emph{HapticNet}}); (b) classification of image input (\emph{VisualNet}); (c) classification of hybrid input (\emph{FusionNet}).}p}
	\label{fig:Net}
\end{figure*}

\noindent\textbf{Haptic Trace/Image Preprocessing}
\revised{As shown in Fig.~\ref{fig:rawsignal}, the haptic acceleration signal usually starts with a short initial impulse signal, when the rigid device initially touches the surface. The following data is a much longer movement signal when the device moves over the surface. As demonstrated by many speech recognition works \cite{Speech:CNN3}\cite{Speech:CNN4}\cite{Speech:CNN5}\cite{Acoustic:music}, converting a 1-D signal into the spectral domain is helpful for the CNN to achieve translation invariance in both temporal domain and frequency domain. Therefore, the 1-D raw acceleration data of the steady-state movement signal is transferred to its spectrogram.

A Hamming window in the time domain is used for enframing haptic signals, where the Hamming window length is set to 500, and the window shift is set to 100. At a sampling rate of 10 kHz, this is equivalent to 50 ms for the window size. Following~\cite{generatingCulbertson}, acceleration segments recorded during unconstrained exploration procedures generally stay stationary in such frame sizes. We select the first 50 low-frequency channels from the spectrogram, which preserved the most of energy from the haptic signal. Finally, the spectrogram is normalized such that the response in each channel has a minimum and maximum value of $0$ and $1$, respectively.

Image inputs are resized to half-size for preprocessing. In this way, most of the texture patterns in an image can be preserved in the $224\times224$ receptive fields window of AlexNet \cite{CNN:Alex} (please refer to \textbf{\emph{VisualNet}} for more detail on our visual classification approach).}

\vspace{0.3cm}
\noindent\textbf{\emph{HapticNet}}
Unlike previous work which applies CNN for haptic surface material recognition, we use a trained FCN to achieve dense prediction for the spectrogram input. FCN replaces the CNN's fully-connected layers by $1 \times 1$ convolutional layers. Without fully-connected layers, FCN produces dense prediction for the input with arbitrary length. Usually adjacent predictions share overlapping receptive field and intermediate features. Since FCN gets rid of the computational redundancy for the overlapping intermediate features, compared to the approach of `CNN + sliding windows', FCN is much more efficient.

Our proposed FCN network for the haptic (acceleration) data, denoted as {\emph{HapticNet}}, takes input of arbitrary length, and the output is a sequence of softmax-vectors of length $L$, $\{\mathbf{y_1}, \cdots, \mathbf{y_L}\}$ representing the categorical probability distributions at different temporal locations (illustrated in Fig. \ref{fig:Net}(a)). The detailed structure of {\emph{HapticNet}} is shown in Table \ref{fig:HapticNet-structure}. The \emph{HapticNet} consists of three normal convolution max-pooling layers, and two $1 \times 1$ convolutional layers which replace the CNN's fully connected layers. The following layer of the final convolutional layer is the softmax layer, which gives categorical probability distributions at every temporal location. The detailed structure and layer configurations of the {\emph{HapticNet}} are shown in Table \ref{fig:HapticNet-structure}.
\begin{table}[htbp]
    \begin{center}
      \begin{tabular}{ | l | c | c |}
        \hline
        layer type & patch size / stride & \specialcell{output\\channel size} \\ \hline \hline
        convolution & 3*3 / 1 & 50 \\ \hline \hline
        pooling & 2*2 / 2 & 50 \\ \hline \hline
        normalization & - & 50 \\ \hline \hline
        convolution & 3*3 / 1 & 100 \\ \hline \hline
        pooling & 2*2 / 2 & 100 \\ \hline \hline
        convolution & 3*3 / 1 & 150 \\ \hline \hline
        pooling & 2*2 / 2 & 150 \\ \hline \hline
        convolution & 3*3 / 1 & 200 \\ \hline \hline
        pooling & 2*2 / 2 & 200 \\ \hline \hline
        convolution & 4*12 / 1 & 400 \\ \hline \hline
        dropout & - & 400 \\ \hline \hline
        convolution & 1*1 / 1 & 250 \\ \hline \hline
        dropout & - & 250 \\ \hline \hline
        convolution & 1*1 / 1 & 69 \\ \hline \hline
        softmax & - & 69 \\ \hline
      \end{tabular}
        \caption {Detailed structure and layer configurations of the proposed \emph{HapticNet}.}
        \label{fig:HapticNet-structure}
    \end{center}
\end{table}


With the predicted output distribution vectors $\mathbf{y}_i$ at each temporal location $i$ by the FCN, the corresponding class labels $c_i$ are obtained by finding the label with maximum probability, i.e.,
\begin{equation} \label{eq:maxclass}
c_i=\arg \max_{c} (\mathbf{y}_i[c]),
\end{equation}
where $\mathbf{y}_i[c]$ represents the $c$-th element of vector $\mathbf{y}_i$, and $c$ enumerates all texture categories. In order to obtain the class label of the entire haptic signal $C$ from multiple predictions $\{c_1, \cdots, c_n\}$, a max-voting procedure that selects the class label with the maximum vote is adapted,
\begin{equation} \label{eq:maxvote}
C=\arg \max_{c} \sum_{i}(\mathds{1}(c_i=c)).
\end{equation}

\vspace{0.3cm}
\noindent\textbf{\emph{VisualNet}} The classification pipeline for image data is similar to the one used for haptic data, i.e., an input image of arbitrary size is fed into a FCN network (denoted as {\emph{VisualNet}}) for outputting categorical probability distributions of size $M\times N$, $\{\mathbf{y_{1, 1}}, \cdots, \mathbf{y_{M, N}}\}$ at every spatial position. As illustrated in Fig. \ref{fig:Net}(b), the class labels $\{c_{1, 1}, \cdots, c_{M, N}\}$ are obtained by applying Eqn. (\ref{eq:maxclass}) for every $\{\mathbf{y_{i, j}}\}$. Then, the class label of the entire image is obtained by applying Eqn. (\ref{eq:maxvote}) accordingly.

The structure of {\emph{VisualNet}} is motivated by AlexNet which was proposed in \cite{CNN:Alex}. \revised{In order to provide dense prediction, the last three fully connected layers are replaced by $1 \times 1$ convolutional layers (to make the representation simple, we denote the three $1 \times 1$ convolutional layers as $\text{FC}_1$, $\text{FC}_2$ and $\text{FC}_3$ respectively). In addition, the number of output features of $\text{FC}_1$, $\text{FC}_2$ and $\text{FC}_3$ are changed accordingly from \{4096, 4096, 1000\} to \{300, 250, 69\}. The numbers of output features of $\text{FC}_1$ and $\text{FC}_2$ are reduced to prevent overfitting.} We use the learned convolutional layers from AlexNet to initialize our corresponding top convolution layers, and random weights are used to initialize the remaining $1\times1$ convolution layers. The overall pipeline for image classification ({\emph{VisualNet}}) is shown in Fig. \ref{fig:Net}(b), which predicts a distribution vector at every spatial location given the image input.

\vspace{0.3cm}
\noindent\textbf{\emph{FusionNet}}
\revised{Suppose that two material surfaces A and B are similar in their visual appearance, but completely different with regard to their haptic perception. In such case, the haptic features will compensate the visual features for better indicating the surface type, and vice versa. Generally, in the scenario where both haptic and image data can be easily captured, a fusion framework can be proposed, where the hybrid input serves for complementing each of the individual information sources, providing richer features for the material surface classification task.

To fuse predictions using both haptic and image data, the \emph{FusionNet} structure is proposed where image and haptic signals are fed to \emph{VisualNet} and \emph{HapticNet} respectively; then haptic/visual features are randomly {sampled} from \emph{HapticNet/VisualNet}'s convolutional feature maps for $K$ times; finally, the sampled haptic features and visual features are concatenated and fed into a $1 \times 1$ convolutional layer with 69 outputs for final predictions. Given multiple predictions like this, the max-voting is applied to obtain the final output class label. The structure of the proposed \emph{FusionNet} is depicted in Fig. \ref{fig:Net}(c).

Training the \emph{FusionNet} end-to-end is non-trivial, due to the described random sampling procedure which is not a common routine for standard deep neural networks. However, note that when input sizes of \emph{FusionNet} are equal to the respective receptive field sizes of \emph{HapticNet/VisualNet}, only one haptic/visual feature can be sampled from \emph{FusionNet}. In such case, the random sampling procedure can be ignored, while \emph{FusionNet} becomes a conventional fusion network. Exploiting such trait, \emph{FusionNet} training can be much simplified. During testing, however, the described \emph{FusionNet} structure still remains. In all, training of \emph{FusionNet} can be simple, while testing of \emph{FusionNet} can be as efficient as our proposed \emph{HapticNet} and \emph{VisualNet}.

When training \emph{FusionNet}, the \emph{FusionNet} weights are initialized from the pretrained weights of \emph{VisualNet} and \emph{HapticNet}. Also during training, a much larger learning rate (10 times to 50 times) can be employed for the final $1\times1$ convolution layer to speed up the training. More details about the training of \emph{FusionNet} can be found in the experimental results section of the paper.}




\section{Experimental Results and Discussions}
\label{sec:experiment}
\noindent\textbf{Dataset}
\revised{The proposed {\emph{HapticNet}}, {\emph{VisualNet}} and \emph{FusionNet} are evaluated using the TUM Haptic Texture dataset \cite{Haptic:TUMdata}, which contains $69$ texture classes and each class consists of 10 sampled haptic traces + 10 images, respectively. For the haptic data, these free-hand sampled traces vary in force and velocity. For the image data, it contains various samples under different lighting conditions.

The dataset is separated into training set and testing set using a ten-fold cross validation. In each fold, the training set contains 9 haptic traces and 9 images for each texture class, and the testing set contains the one remaining haptic data trace and image in each class. The training set is used to train the proposed network, and the testing set is used to evaluate the performance of the network and the following max-voting scheme. Before the ten-fold cross validation, a train/validation/test split is adapted for networks hyper-parameters tuning.}

\vspace{0.3cm}
\noindent\textbf{Training Details}
\revised{Our implementation is based on the Caffe deep learning framework \cite{CNN:Caffe}, using a computer with 8GB RAM and Nvidia GPU GTX-860M. The training of {\emph{HapticNet}}, {\emph{VisualNet}} and \emph{FusionNet} is conducted by applying Adam \cite{CNN:Adam}, an gradient-based stochastic optimization algorithm. For training each network, the learning rates are initialized to the base learning rate $\lambda_0$, then dropping by a constant factor $\gamma$ every $k$ iterations. Different choices of $\lambda_0$, $\gamma$ and $k$, as well as the total training iterations $N$ for each network is depicted in Table \ref{table:allParams}. The $L2$ weight decay is set to $5\times10^{-4}$. Parameters of Adam are set to the default value proposed by \cite{CNN:Adam}, specifically $\beta_1=0.9$, $\beta_2=0.999$ and $\epsilon=10^{-8}$.

During the testing stage of \emph{HapticNet}, the preprocessing procedures described in Section~\ref{subsection:surface_classification} are first applied to the haptic tracks. The preprocessed data are then input to \emph{HapticNet}. In the training stage, however, the haptic spectrograms are subsampled using a fixed length, then being fed into the network for training. By enforcing subsamples to be consistent in length when training \emph{HapticNet}, we are able to increase the mini-batches size and thus speed up training. The size of the training subsample and mini-batch is depicted in Table~\ref{table:allParams}.

Similarly, in the testing stage of \emph{VisualNet}, preprocessed image is fed into the network for prediction. In the training stage, the preprocessed images are first subsampled into fixed-size patches, then fed into the network for training. Data augmentation using random rotation is also applied during \emph{VisualNet} training. The size of the training subsample and minibatch is also depicted in Table~\ref{table:allParams}.

To train \emph{FusionNet}, we restrict the size of the haptic and image input to be small enough, such that \emph{HapticNet} will generate a single prediction for haptic input and \emph{VisualNet} will only generate a single prediction for image input (as also described in \emph{FusionNet} of Section~\ref{subsection:surface_classification}). The input size and mini-batch settings for \emph{FusionNet} are depicted in Table \ref{table:allParams}. During testing, we set the randomly sampled number $K$ described in \emph{FusionNet} of Section~\ref{subsection:surface_classification} to 1000. However, we do not observe other numbers to induce a significant influence on the classification performance.}

\begin{table}[htbp]
    \begin{center}
        \begin{tabular}{l*{6}{c}r}
\hline\hline
							&$\lambda_0$			&$\gamma$ &$k$	  & $N$ 	&\specialcell{input\\size} & \specialcell{batch-\\size} \\
        \hline
        {HapticNet}    & $10^{-4}$ 			& $0.3$   & $4 \times 10^{4}$ & $10^{5}$  & $50 * 300$				& $10$ 	\\
        \hline
        {VisualNet}    & $3 \times 10^{-5}$ 	& $0.75$  & $4 \times 10^{4}$ & $10^{5}$  & $384 * 384$				& $2$ 	\\
        \hline
        {FusionNet}    & $10^{-6}$ 			& $0.1$   & $4 \times 10^{6}$ & $10^{5}$  & \specialcell{$50*192$ \\ \& \\ $224*224$} & $1$	\\
        \hline\hline
        \end{tabular}
        \caption {\revised{The hyper-parameters for training {\emph{HapticNet}}, {\emph{VisualNet}} and \emph{FusionNet}.}}
        \label{table:allParams}
    \end{center}
\end{table}

\begin{table}[h!]
    \begin{center}
        \begin{tabular}{l*{3}{c}r}
        \hline\hline
        \emph{Haptic Classification}                 & \emph{Fragment} & \emph{Max voting}  \\
        \hline
        MFCC + GMM~\cite{Haptic:TUMdata}                                   &    --    & 80.23\%   \\
        \specialcell{Modified MFCC Decreasing \\+ Naive Bayes~\cite{Haptic:Recognition1}}  &    --    & 89\%      \\
        ACNN~\cite{Haptic:Mengqi}	& 81.8\% & -- \\
        \textbf{{HapticNet}}                    & \textbf{85.3\%} & \textbf{91.0\%} \\
        \hline\hline

        \emph{Visual Classification}                & \emph{Fragment} & \emph{Max voting}  \\
        \hline
        TCNN~\cite{TextureImage:Using filter banks in CNN} & {87.1\%} & -- \\
        VGG-M-FV~\cite{TextureImage:Deep filter banks}  & 73.7\% & --  \\
        AlexNet-FV~\cite{TextureImage:Deep filter banks}  & 78.7\% & --  \\
        {VisualNet}  & 85.6\% & 93.3\%  \\
        \textbf{VisualNet-TCNN}   & \textbf{87.1\%} & \textbf{95.5\%}  \\
        \hline\hline

        \emph{Hybrid Classification}                & \emph{Fragment} & \emph{Max voting}  \\
        \hline
        {FusionNet-$\text{FC}_3$}                       & 95.0\% & 98.1\% \\
        {FusionNet-$\text{FC}_2$}                      & 96.2\% & 98.4\% \\
        {{FusionNet-$\text{FC}_2$-TCNN}}     & {96.6\%} & 98.4\% \\
        \textbf{{FusionNet-$\text{FC}_3$-TCNN}} 		& \textbf{96.6\%} & \textbf{98.8\%} \\
        \hline\hline

        \end{tabular}
        \caption {\revised{The surface classification results using haptic data and / or visual data.}}
        \label{table:allResult}
    \end{center}
\end{table}

\vspace{0.3cm}
\noindent\textbf{\emph{HapticNet}} The classification accuracy is shown in the \emph{Haptic Classification} part of Table \ref{table:allResult}. Specifically, we compare our results with several existing methods \cite{Haptic:TUMdata}\cite{Haptic:Recognition1}\cite{Haptic:Mengqi}. \cite{Haptic:TUMdata} proposes to combine MFCC features with a Gaussian Mixture Model (GMM) for movement phase recognition, which is denoted as `MFCC + GMM' here. The following work \cite{Haptic:Recognition1} carefully discusses variant features for representing the movement signal, and variant discriminative models that are proposed for giving predictions. Considering \cite{Haptic:TUMdata} uses ``averaging feature'' to test the prediction accuracy, which performs an underlying model averaging, the obtained results are listed in the max-voting column. \cite{Haptic:Mengqi} is the first work which uses CNN to classify the raw haptic data. By comparison, \emph{HapticNet} achieves superior performance with the fragment accuracy be 85.3\% and max-voting accuracy be 91.0\%, using the ten-fold cross validation measurement.

%
%

\vspace{0.3cm}
\noindent\textbf{\emph{VisualNet}}
\revised{The classification accuracy of the concerned scheme is shown in the \emph{Visual Classification} part of Table \ref{table:allResult}. Specifically, we compare our approach \emph{VisualNet} with the T-CNN-3 (denoted as TCNN) in \cite{TextureImage:Using filter banks in CNN} and the descriptor (denoted as \emph{VGG-M-FV}) in \cite{TextureImage:Deep filter banks}, as well as the descriptor by replacing VGG-M network with AlexNet (denoted as \emph{AlexNet-FV}). We observe that \emph{VisualNet} achieves comparable fragment accuracy as \emph{TCNN} \cite{TextureImage:Using filter banks in CNN}, and higher accuracy than \emph{VGG-M-FV} \cite{TextureImage:Deep filter banks} and \emph{AlexNet-FV} \cite{TextureImage:Deep filter banks}. To demonstrate that the \emph{VisualNet} framework is easily adaptable for other neural network design, we adapt the \emph{TCNN} into a fully convolutional version, which is denoted as \emph{VisualNet-TCNN}. \emph{VisualNet-TCNN} is achieved by reserving the first three convolutional layers of \emph{VisualNet}, followed by an average pooling layer (which is similar to \emph{TCNN}) and three $1\times 1$ convolutional layers. \emph{VisualNet-TCNN} achieves highest accuracy for classifying images of surface materials.

Additionally, we test \emph{VisualNet} and \emph{VisualNet-TCNN} on other texture image datasets, including Kylberg Texture Dataset \cite{TextureImage:Kylberg} and KTH-TIPS-2b \cite{TextureImage:KTH}. Our experimental results are reported in Table \ref{table:Texture-dataset-Result}. Note that the train/test split of Kylberg Texture Dataset and KTH-TIPS-2b follows the paper \cite{TextureImage:Using filter banks in CNN}. On Kylberg Texture Dataset, the T-CNN-3 (denoted as TCNN) in \cite{TextureImage:Using filter banks in CNN} and the D$_{8}$ descriptor \cite{TextureImage:D8} are compared with our approach. From comparison, \emph{VisualNet} achieves the best performance, with very similar yet slightly higher accuracy than \emph{Visual-TCNN}. It is consistent with the conclusion in \cite{TextureImage:Using filter banks in CNN}, which suggests AlexNet slightly outperforms TCNN on Kylberg dataset. On KTH-TIPS-2b dataset, TCNN and the VGG-M-FV descriptor \cite{TextureImage:Deep filter banks} are compared with our approach. From the comparison we can conclude that \emph{VisualNet} achieves competitive performance. We notice that max-voting obtains only a slight performance gain for the fragment accuracy. This is due to testing image in KTH-TIPS-2b dataset is resized to $256\times256$ -- very small size for the receptive field sizes of \emph{VisualNet} and \emph{VisualNet-TCNN}.}

\begin{table}[h!]
    \begin{center}
        \begin{tabular}{l*{3}{c}r}
        \hline\hline
        \emph{Kylberg Texture Dataset}                & \emph{Fragment} & \emph{Max voting}  \\\hline
        \textbf{{VisualNet}}  & \textbf{96.9\%} & \textbf{97.8\%}  \\
        {VisualNet-TCNN}  & 96.3\% & 96.9\%  \\
        TCNN \cite{TextureImage:Using filter banks in CNN}			  & 96.00\% & -- 	 \\
        D$_{8}$ \cite{TextureImage:D8}		  & 82.0\% & --      \\
        \hline
        \emph{KTH-TIPS-2b Dataset}              			  & \emph{Fragment} & \emph{Max voting}  \\\hline
        {VisualNet}  & 72.0\% & 72.1\%  \\
        \textbf{{VisualNet-TCNN}} & \textbf{72.4\%} & \textbf{72.4\%}  \\
        TCNN \cite{TextureImage:Using filter banks in CNN}			  & 72.36\% & -- 	 \\
        VGG-M-FV \cite{TextureImage:Deep filter banks}     & 73.3\% &  --     \\
        \hline\hline
		\end{tabular}
        \caption {\revised{The \emph{VisualNet} classification results Kylberg Texture Dataset and KTH-TIPS-2b dataset.}}
        \label{table:Texture-dataset-Result}
    \end{center}
\end{table}



\vspace{0.3cm}
\noindent\textbf{\emph{FusionNet}}
\revised{For the experiments of classifying hybrid (haptic+visual) input, we first denote the last three $1 \times 1$ convolutional layers of \emph{HapticNet/VisualNet} as $\text{FC}_1$, $\text{FC}_2$ and $\text{FC}_3$ respectively. The \emph{FusionNet} is tested with two different settings: haptic features and visual features from either $\text{FC}_3$ layers or $\text{FC}_2$ layers are fused (the resulting network are denoted as \emph{FusionNet}-$FC_3$ and \emph{FusionNet}-$FC_2$ respectively in the \emph{Fusion Classification} part of Table \ref{table:allResult}). As we expected, the accuracy of haptic predictions and image predictions is significantly boosted with the fusion framework. By replacing the visual submodule in \emph{FusionNet} with \emph{VisualNet-TCNN}, we have \emph{FusionNet}-$FC_3$-\emph{TCNN} and \emph{FusionNet}-$FC_2$-\emph{TCNN}, which achieve comparable and highest accuracy for classifying surface material.}


%

\begin{figure*}[htbp]
	\centering
	\subfigure[HapticNet]{\includegraphics[width = \colw\linewidth]{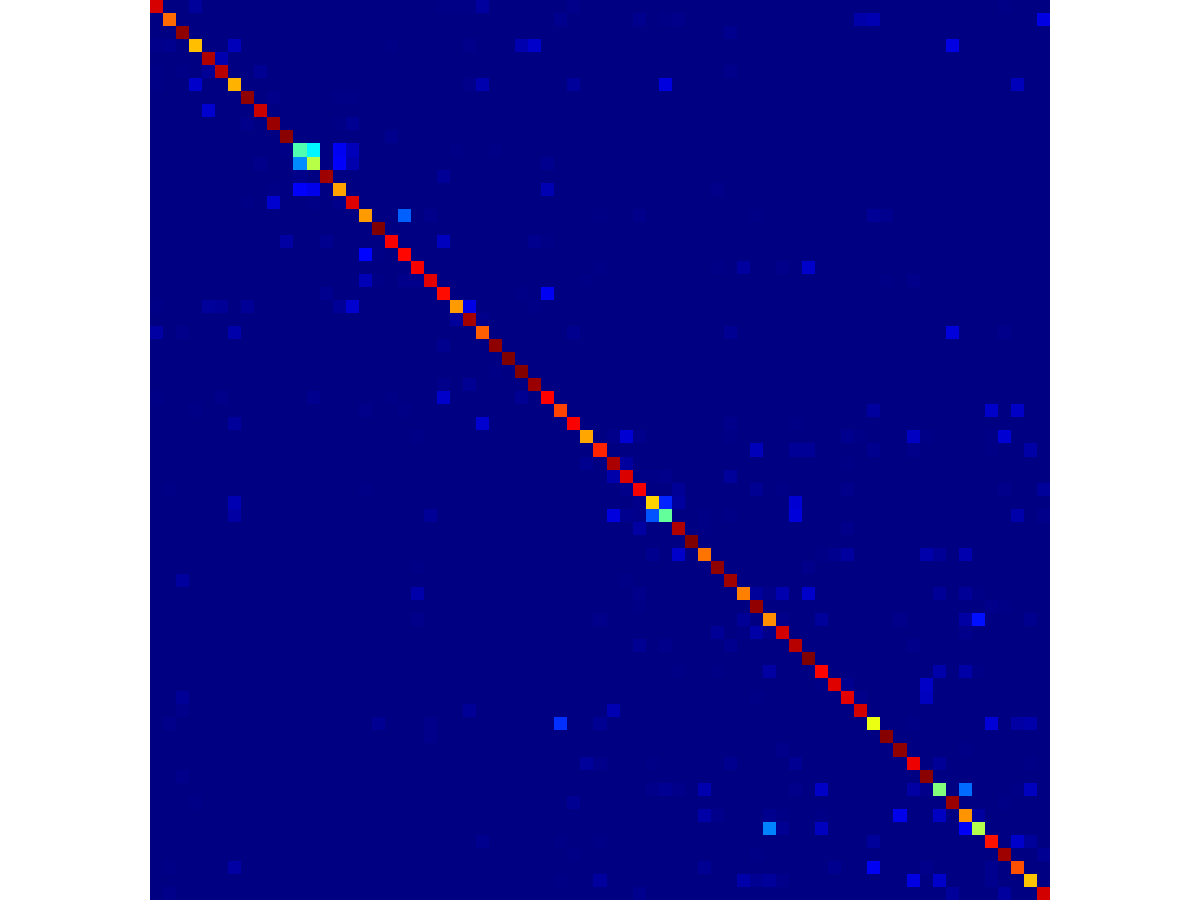}}
	\subfigure[VisualNet]{\includegraphics[width = \colw\linewidth]{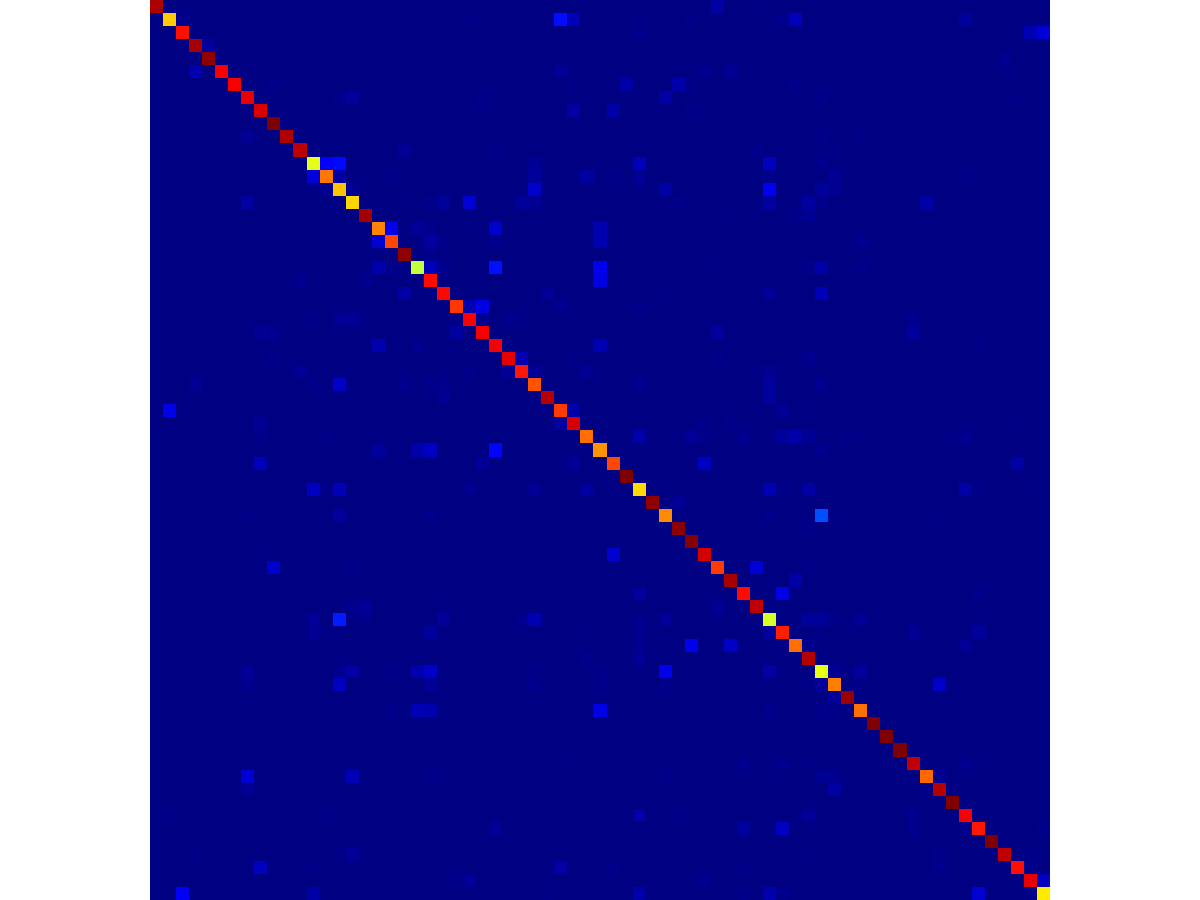}}
	\subfigure[FusionNet-$\text{FC}_2$]{\includegraphics[width = \colw\linewidth]{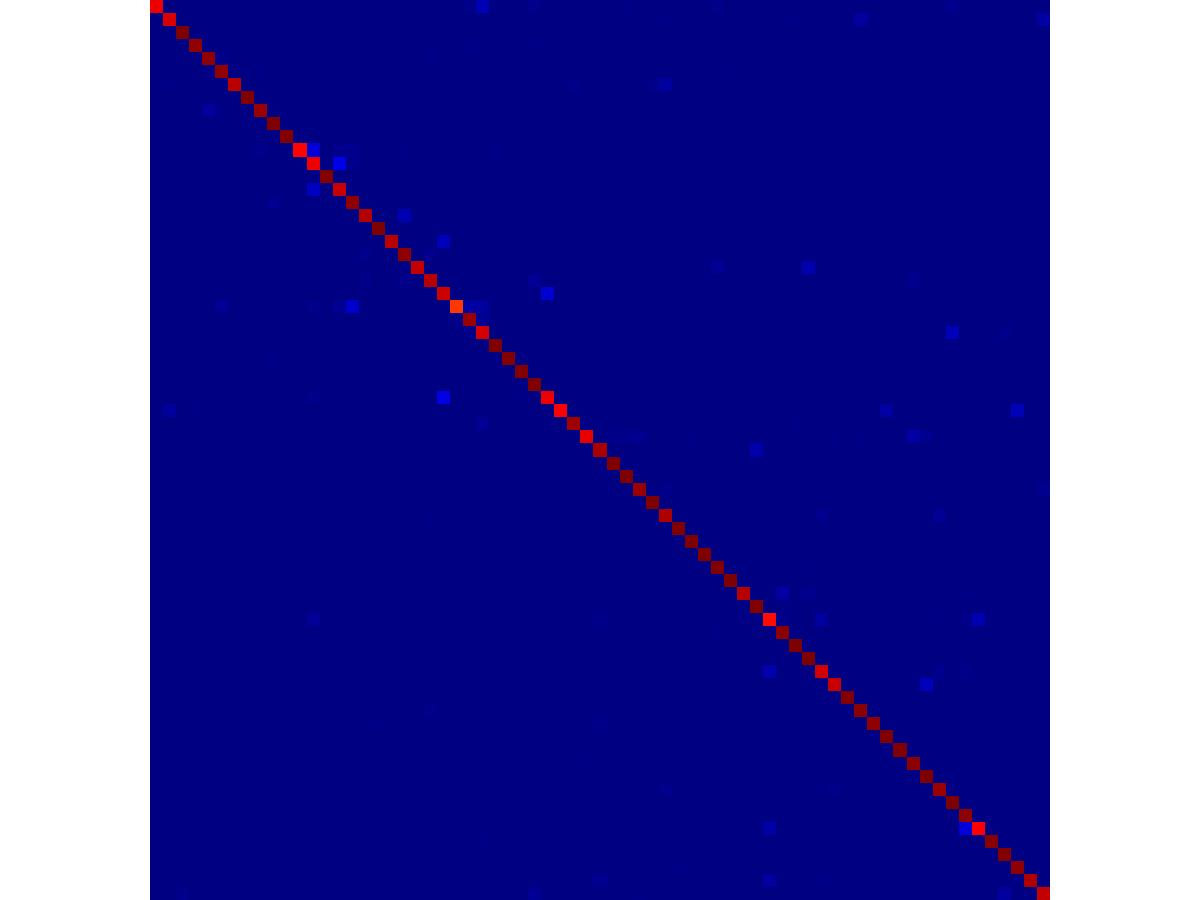}}
	\caption{\revised{The fragment classification confusion matrix (10-fold averages) of (a) \emph{HapticNet}; (b) \emph{VisualNet}; (c) \emph{FusionNet-$\text{FC}_2$}.}}
	\label{fig:confusion-matrices-track}
\end{figure*}

\begin{figure*}[htbp]
	\centering
	\subfigure[Type 12 VS 13]{
			\begin{minipage}[b]{\colw\linewidth}
	    		\includegraphics[width=\figw\textwidth]{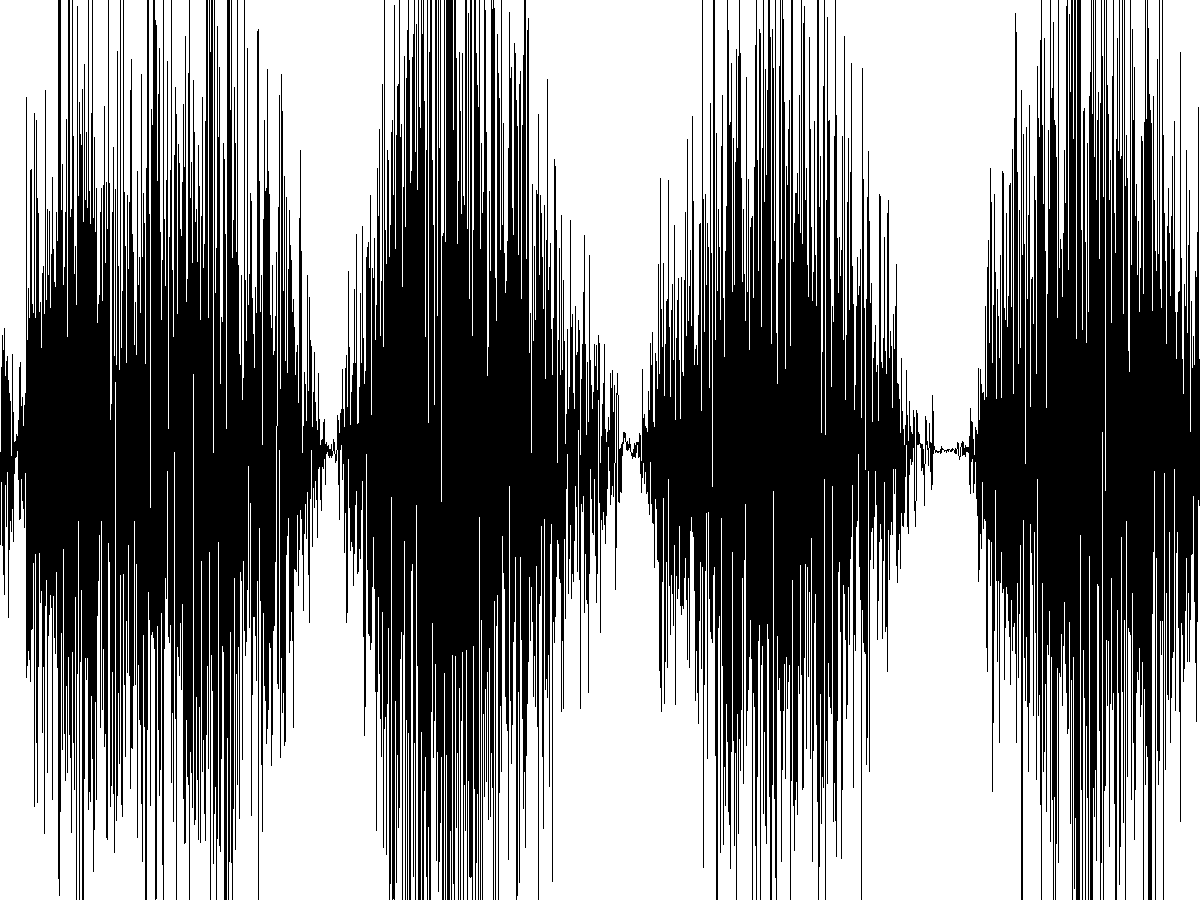}
	    		\begin{overpic}[width=\figw\textwidth]{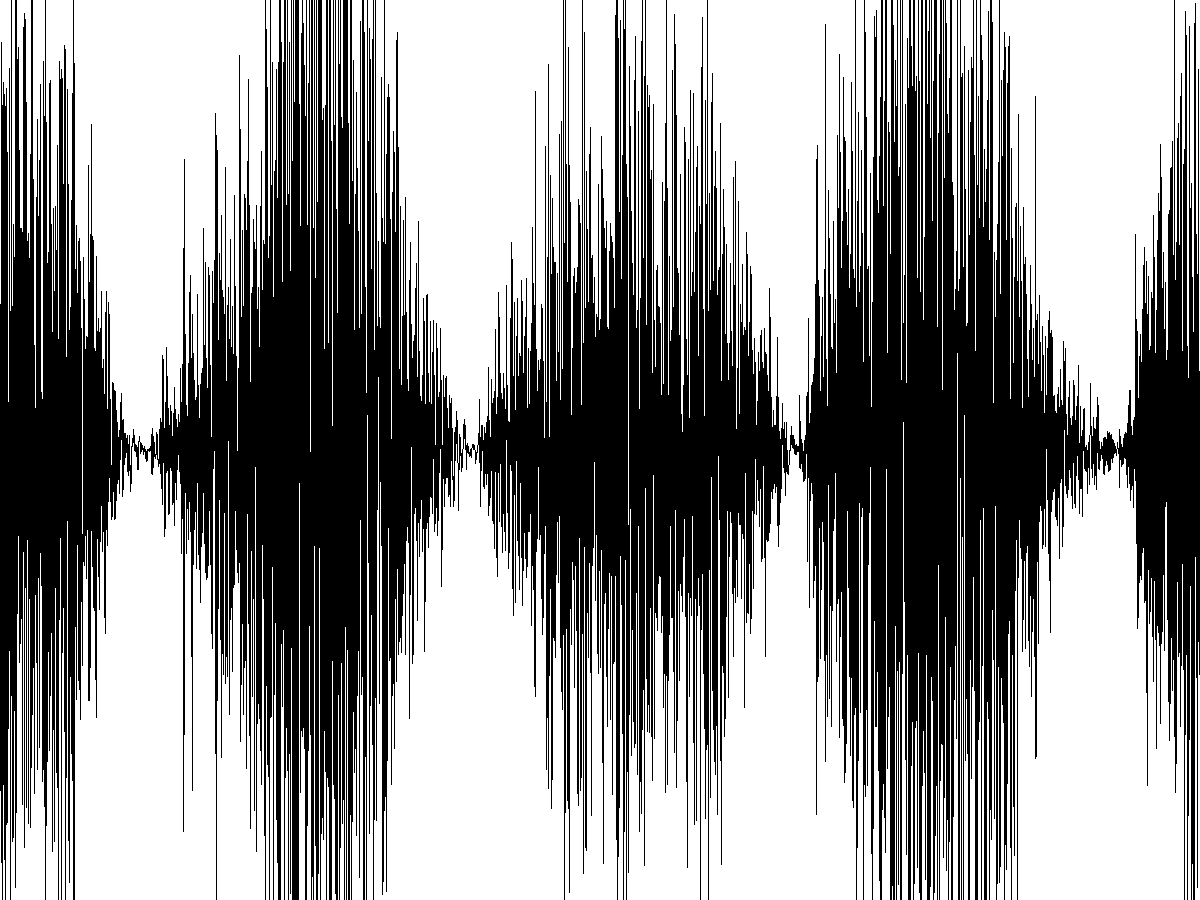}\end{overpic}\\
	    		\begin{overpic}[width=\figw\textwidth]{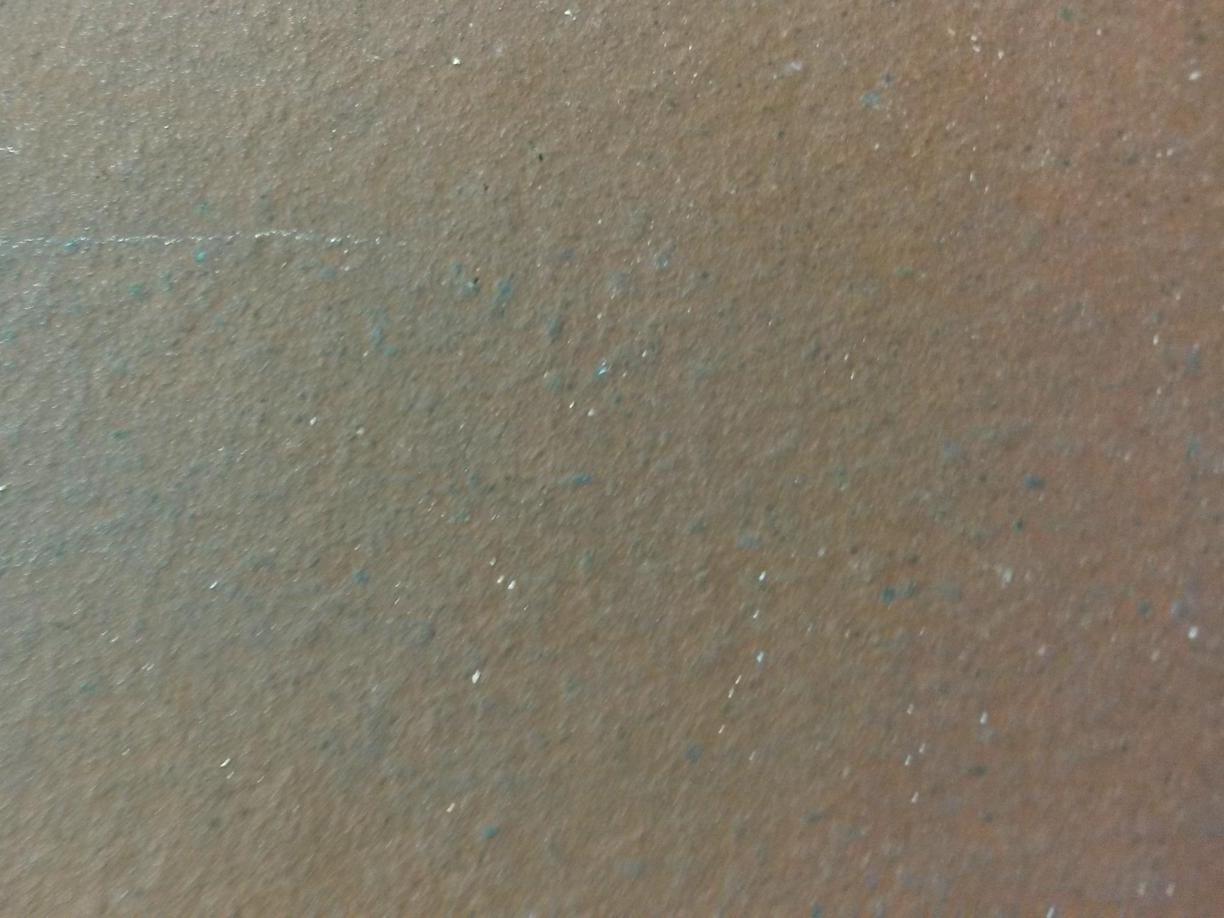}\end{overpic}
	    		\begin{overpic}[width=\figw\textwidth]{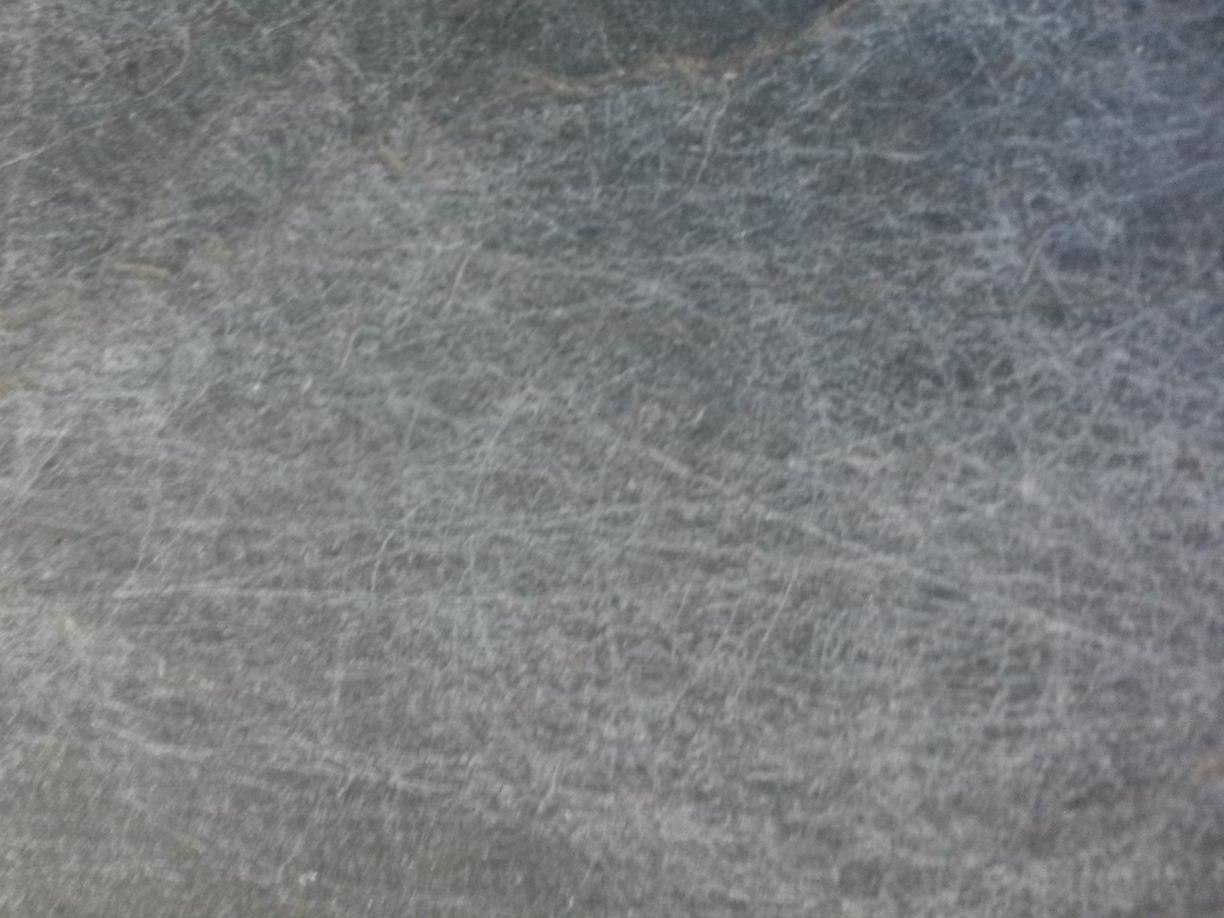}\end{overpic}
	    	\end{minipage}	
	}
	\label{subfig:samples-HapticNetWrong:12-13}
	\subfigure[Type 39 VS 40]{
			\begin{minipage}[b]{\colw\linewidth}
	    		\includegraphics[width=\figw\textwidth]{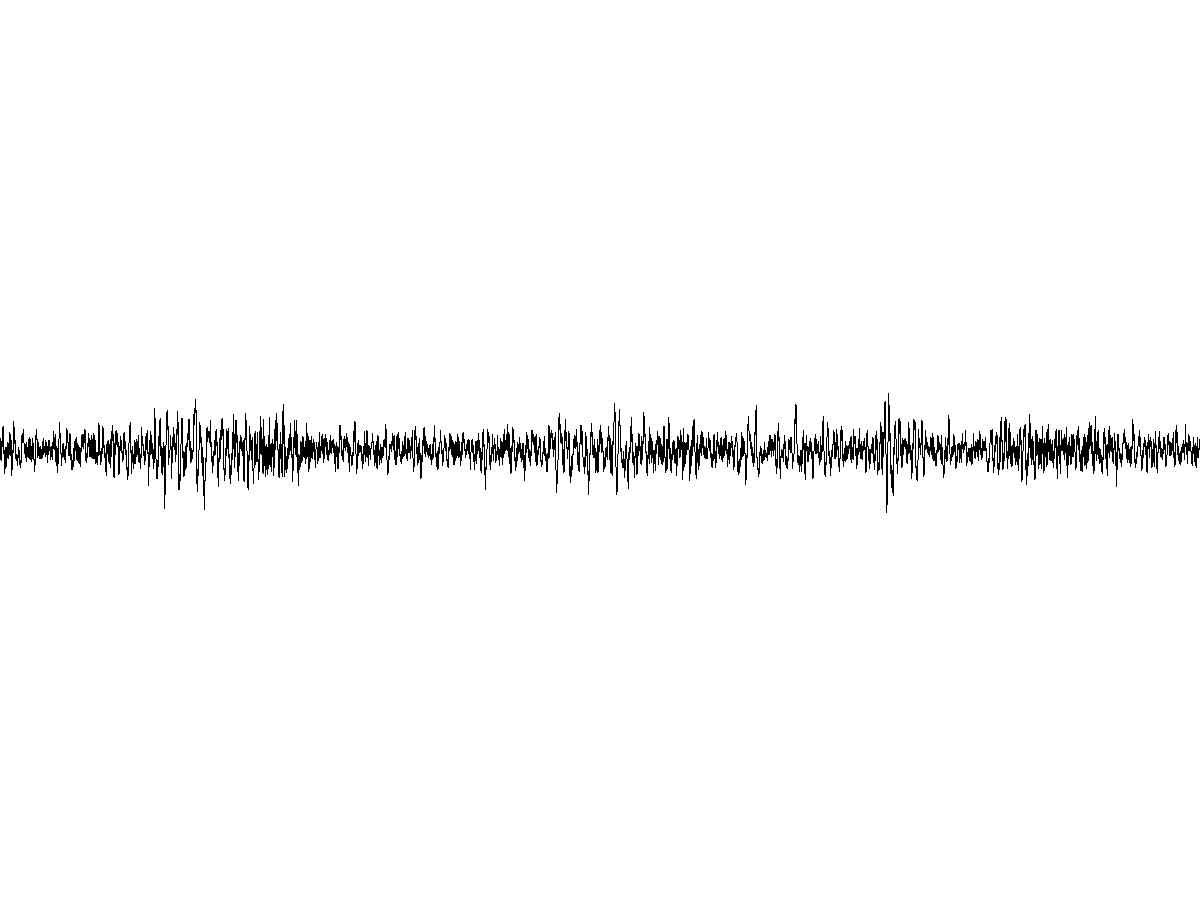}
	    		\begin{overpic}[width=\figw\textwidth]{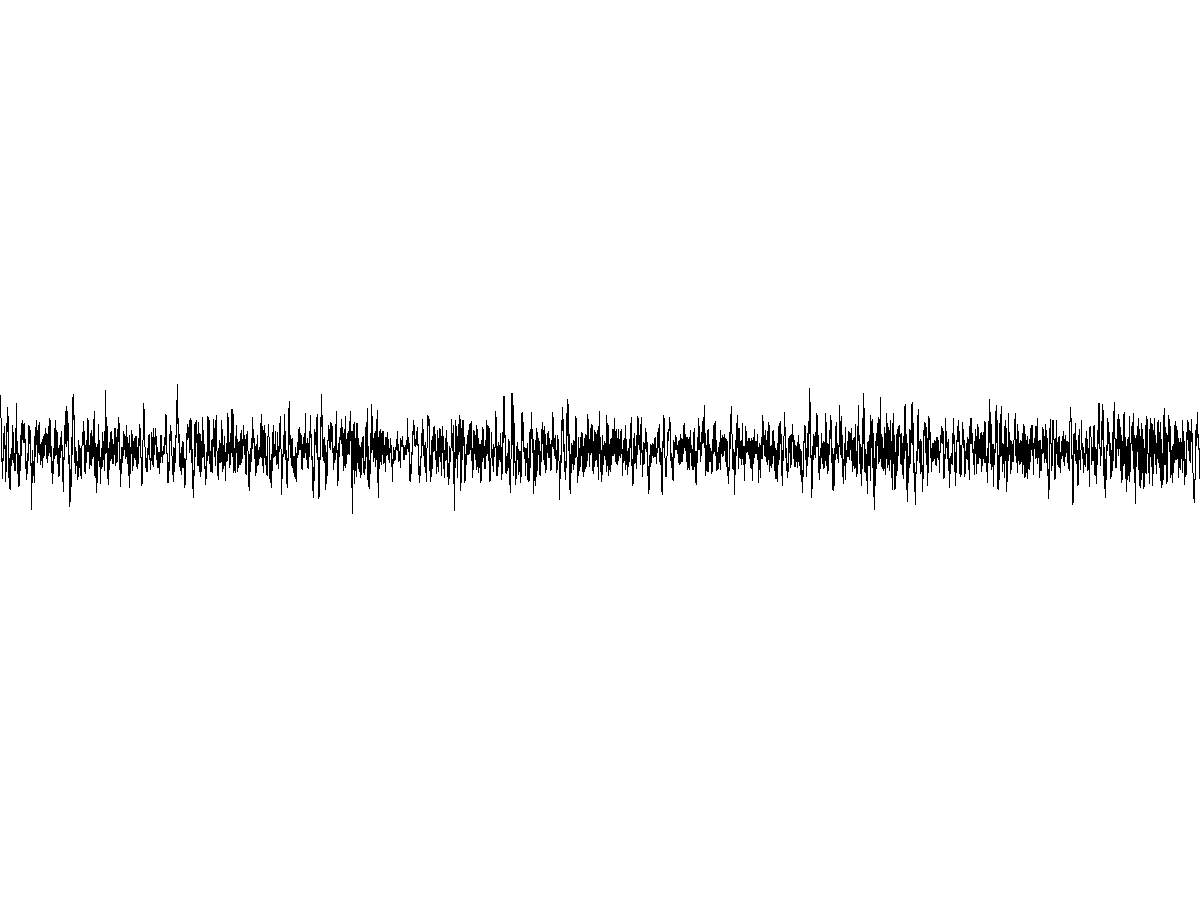}\end{overpic}\\
	    		\begin{overpic}[width=\figw\textwidth]{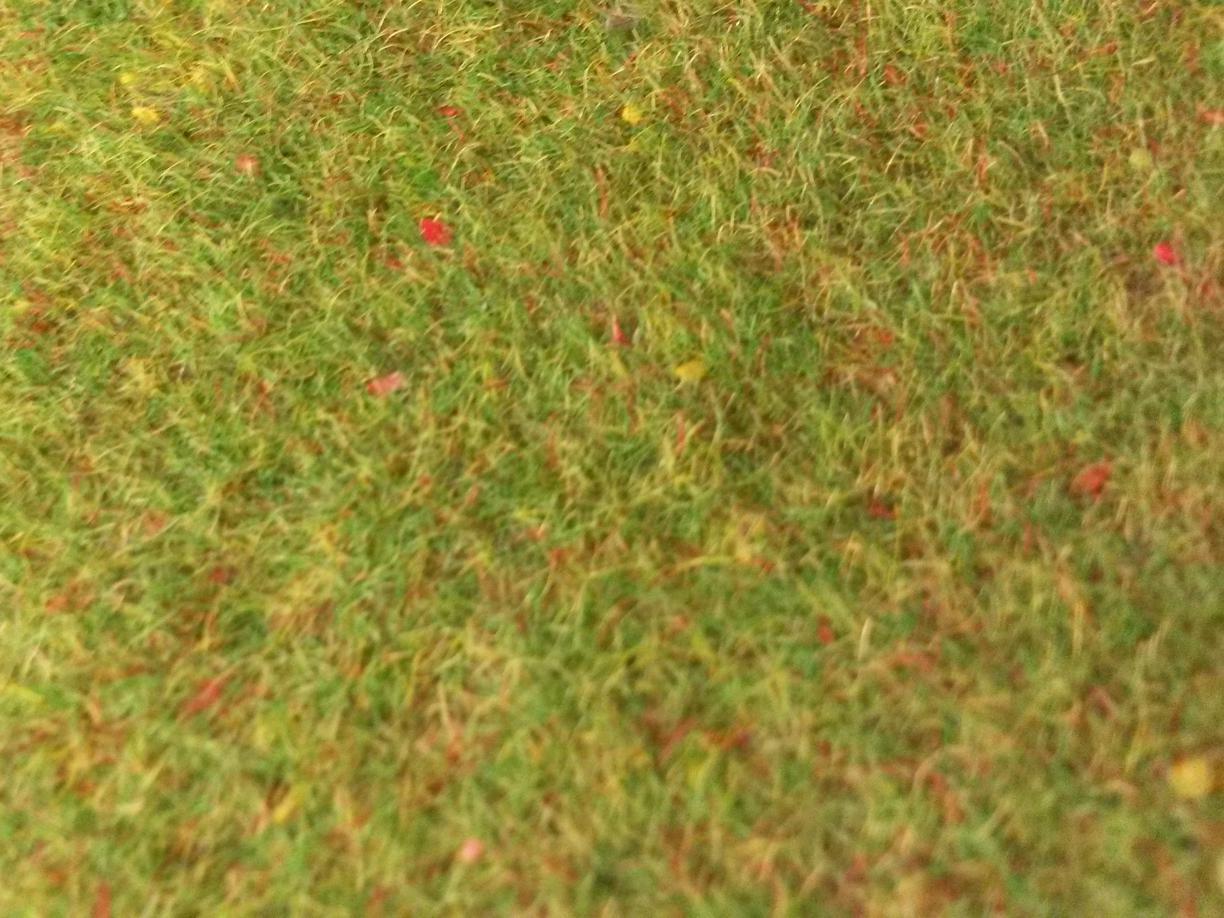}\end{overpic}
	    		\begin{overpic}[width=\figw\textwidth]{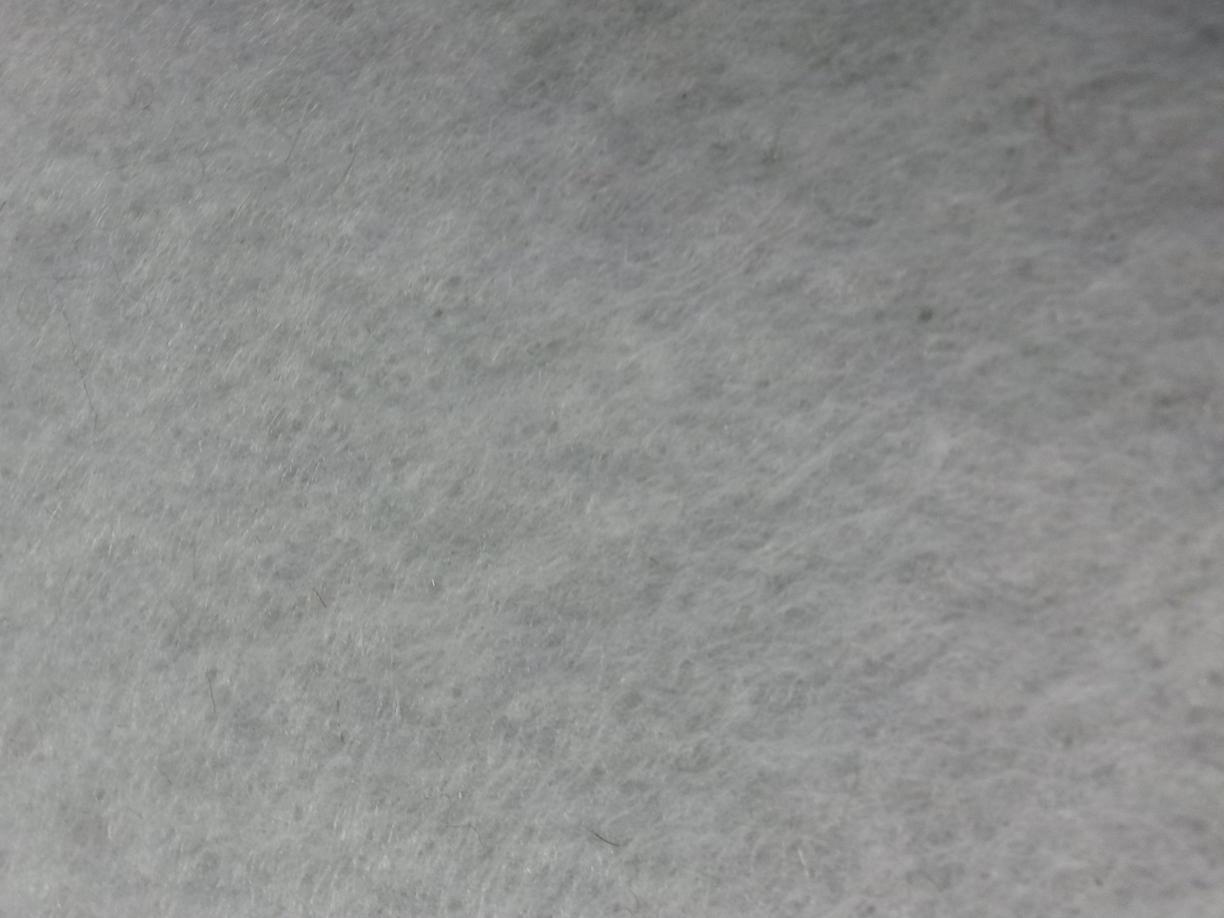}\end{overpic}
	    	\end{minipage}	
	}
	\label{subfig:samples-HapticNetWrong:39-40}
	\subfigure[Type 48 VS 64]{
			\begin{minipage}[b]{\colw\linewidth}
	    		\includegraphics[width=\figw\textwidth]{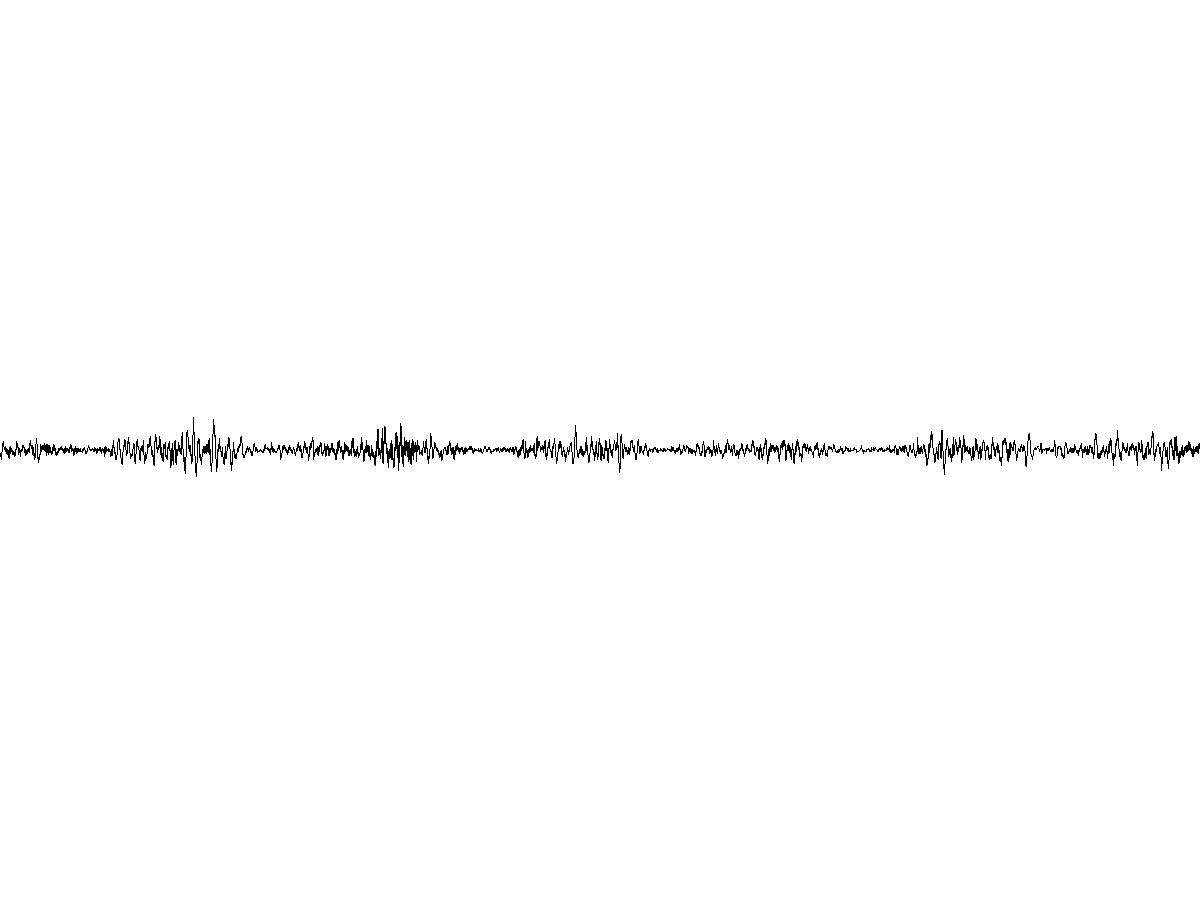}
	    		\begin{overpic}[width=\figw\textwidth]{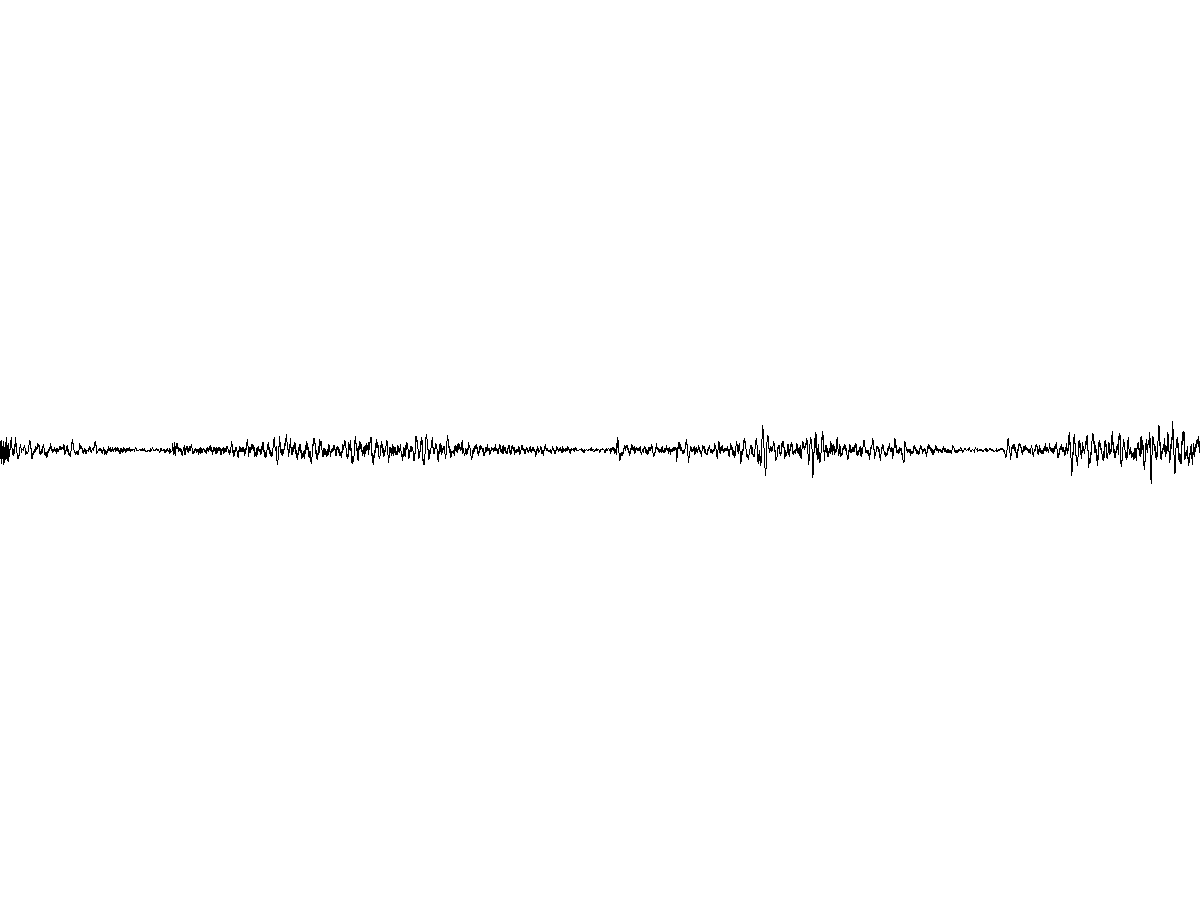}\end{overpic}\\
	    		\begin{overpic}[width=\figw\textwidth]{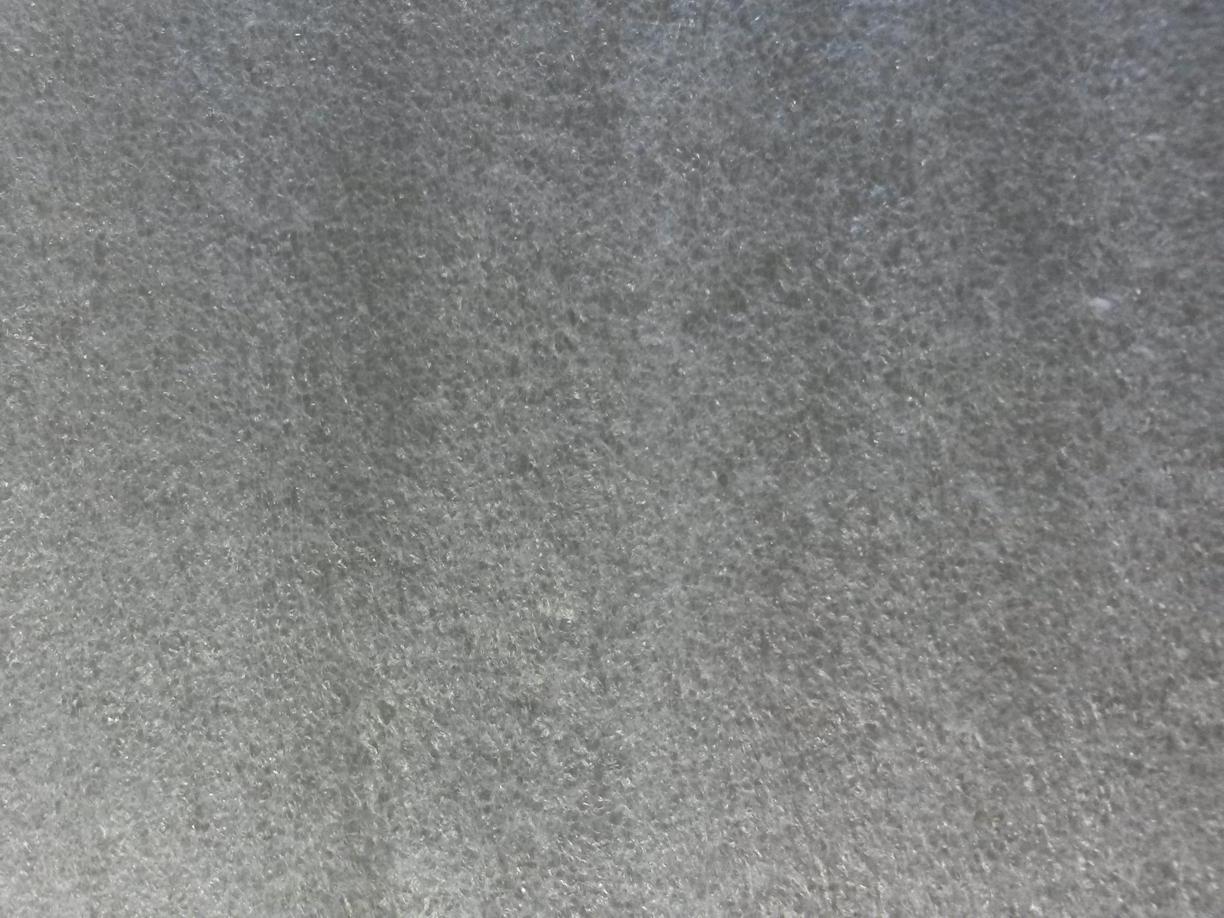}\end{overpic}
	    		\begin{overpic}[width=\figw\textwidth]{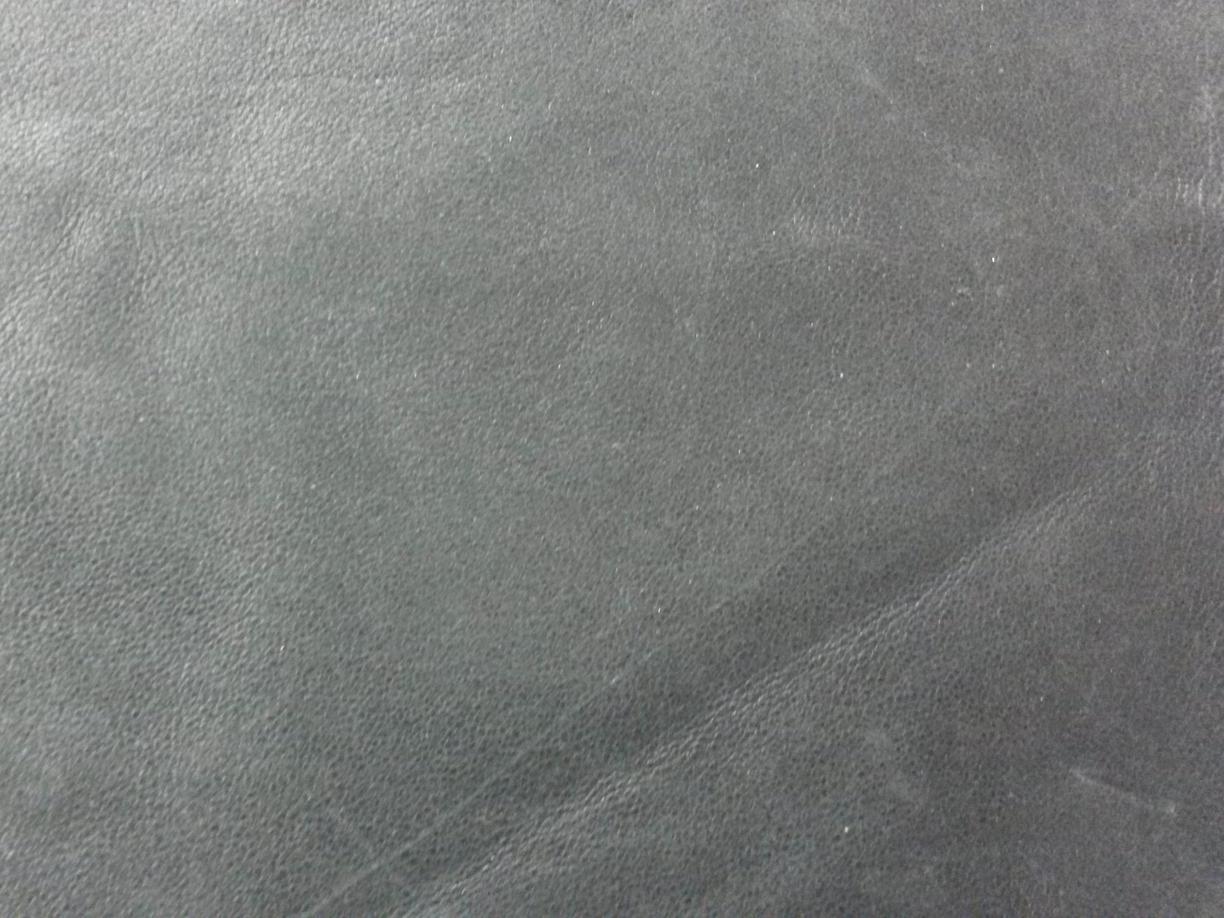}\end{overpic}
	    	\end{minipage}	
	}
	\label{subfig:samples-HapticNetWrong:48-64}
	\subfigure[Type 13 VS 14]{
		\begin{minipage}[b]{\colw\linewidth}
    		\includegraphics[width=\figw\textwidth]{Experiment_figures/material/13/haptic_5.png}
    		\begin{overpic}[width=\figw\textwidth]{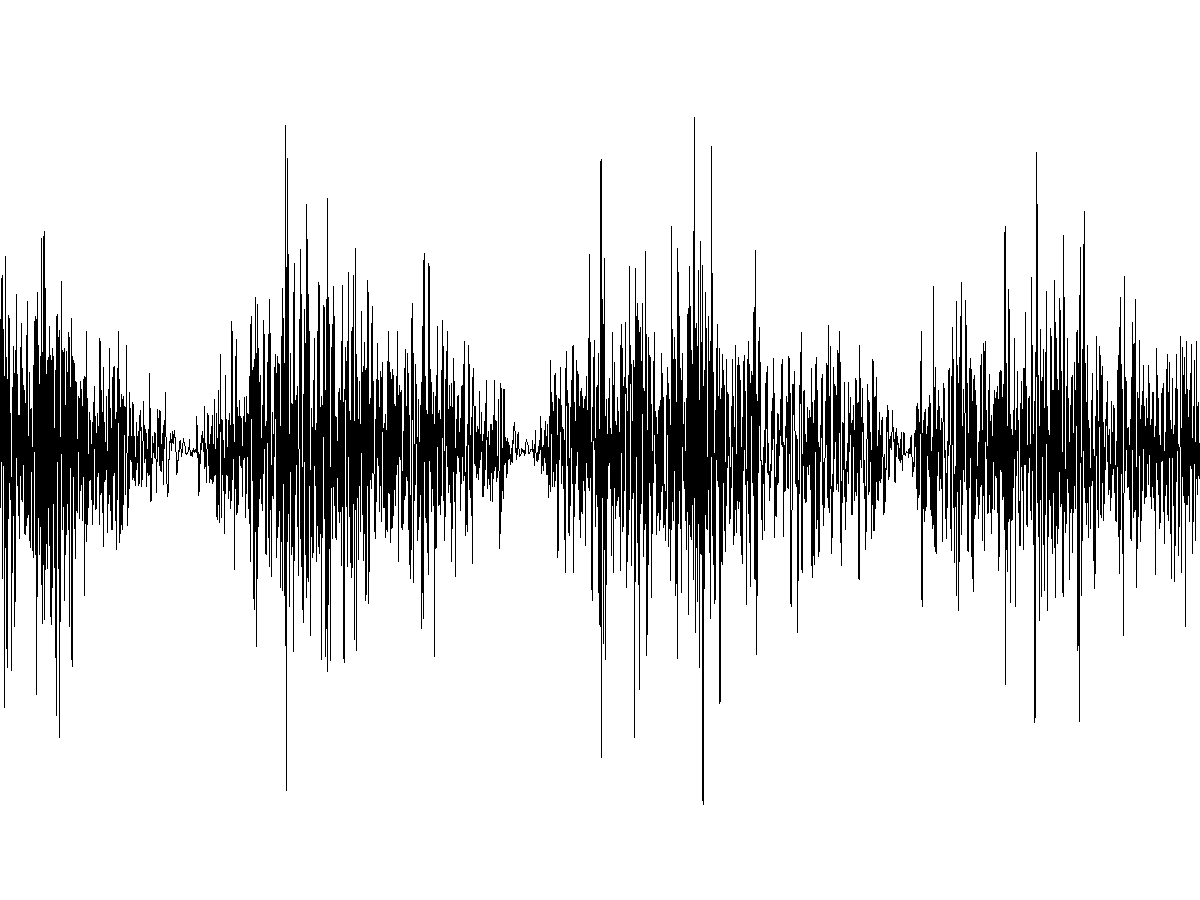}\end{overpic}\\
    		\begin{overpic}[width=\figw\textwidth]{Experiment_figures/material/13/1_rezied.jpg}\end{overpic}
    		\begin{overpic}[width=\figw\textwidth]{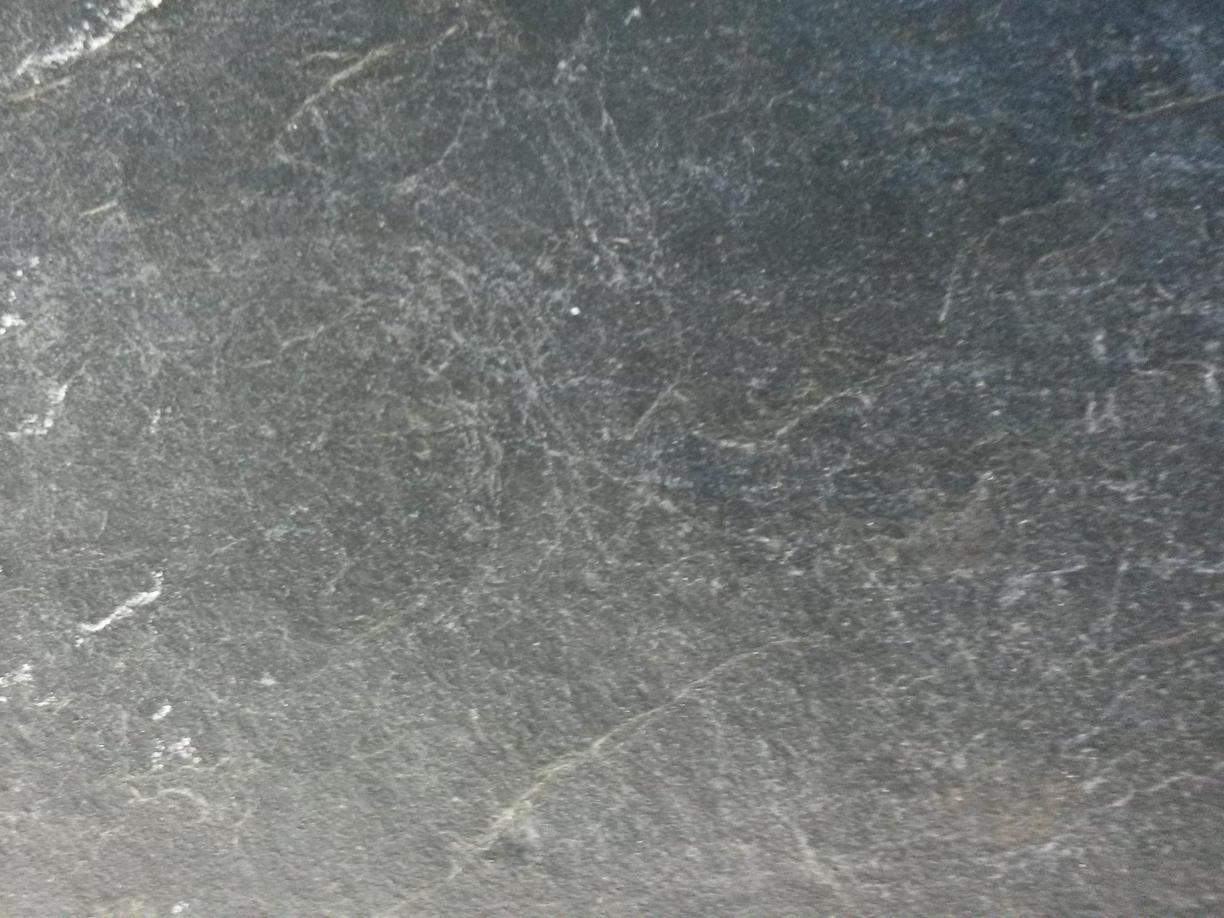}\end{overpic}
    	\end{minipage}
	}
	\label{subfig:samples-VisualNetWrong:13-14}
	\subfigure[Type 18 VS 19]{
			\begin{minipage}[b]{\colw\linewidth}
	    		\includegraphics[width=\figw\textwidth]{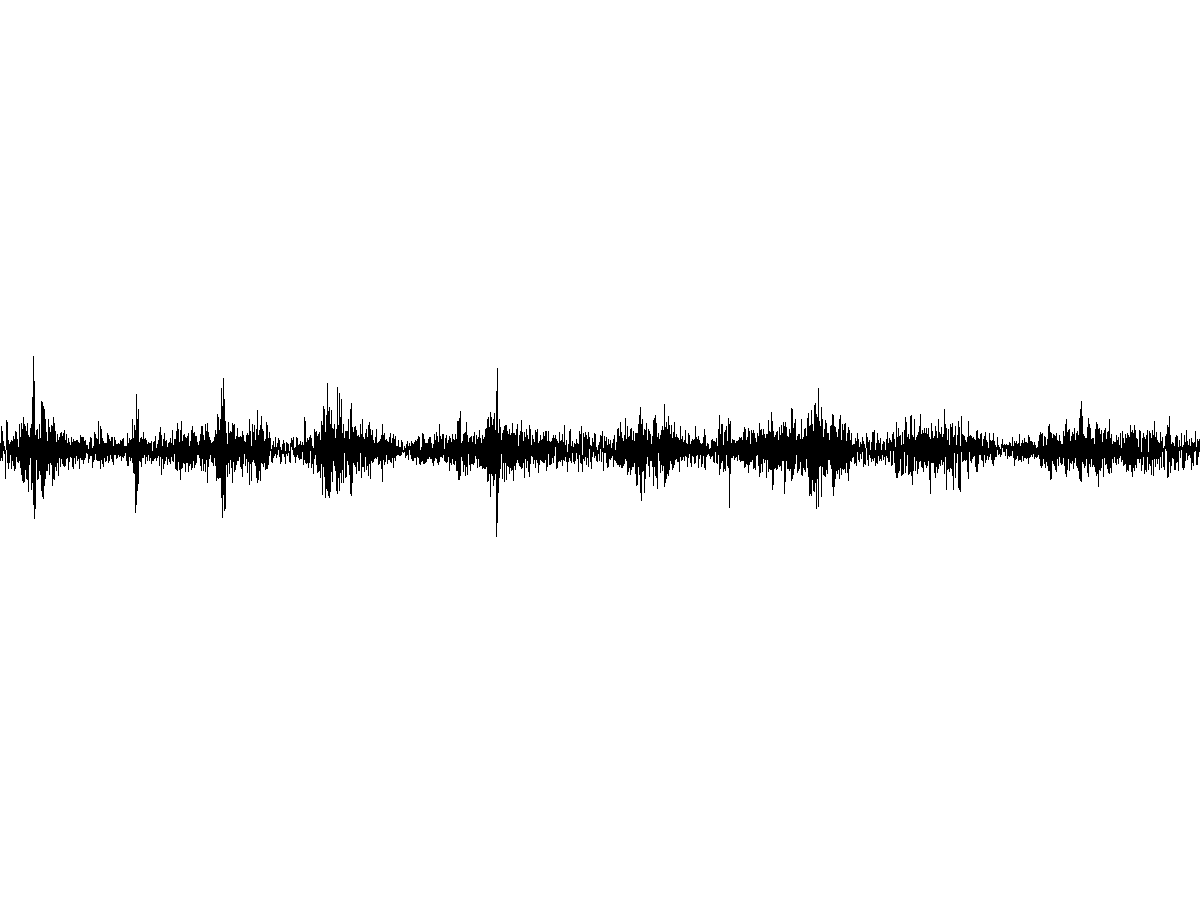}
	    		\begin{overpic}[width=\figw\textwidth]{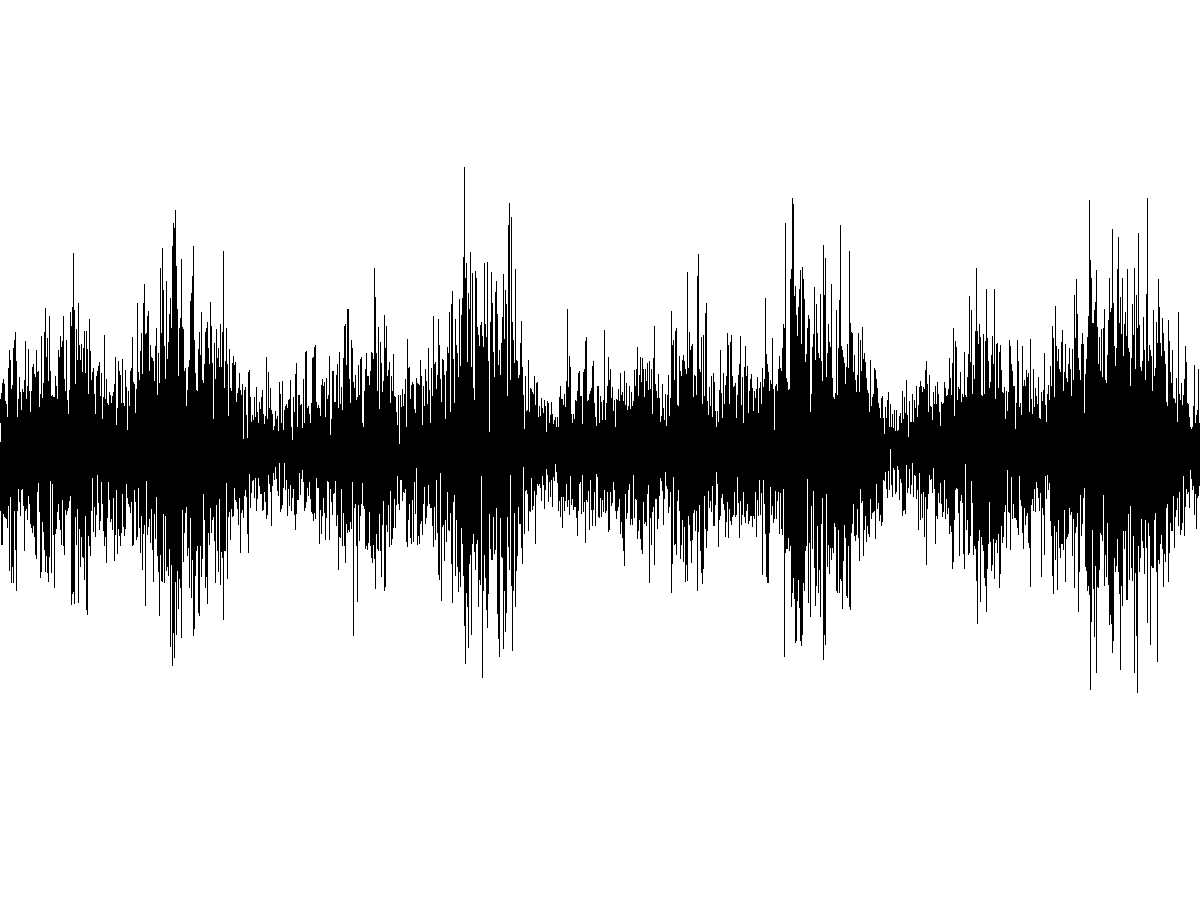}\end{overpic}\\
	    		\begin{overpic}[width=\figw\textwidth]{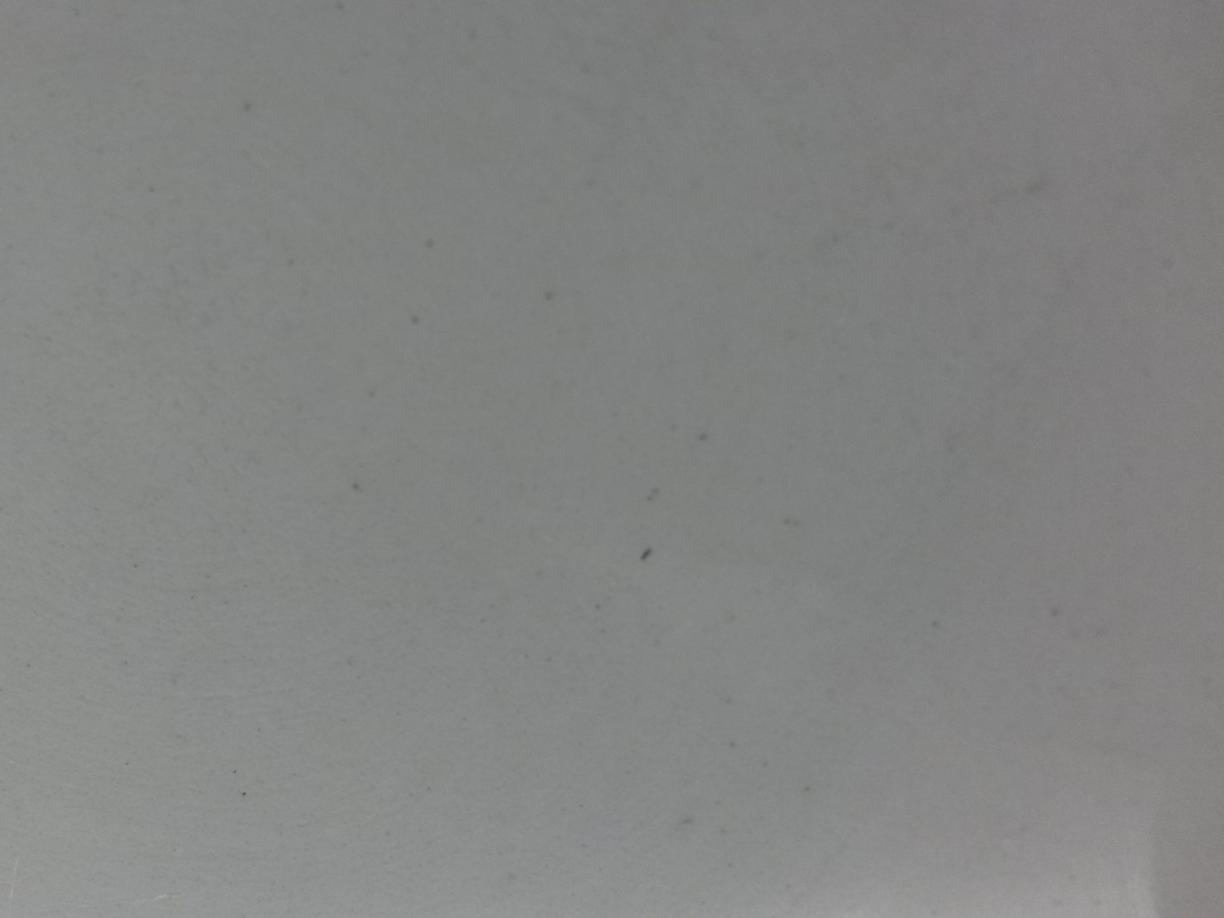}\end{overpic}
	    		\begin{overpic}[width=\figw\textwidth]{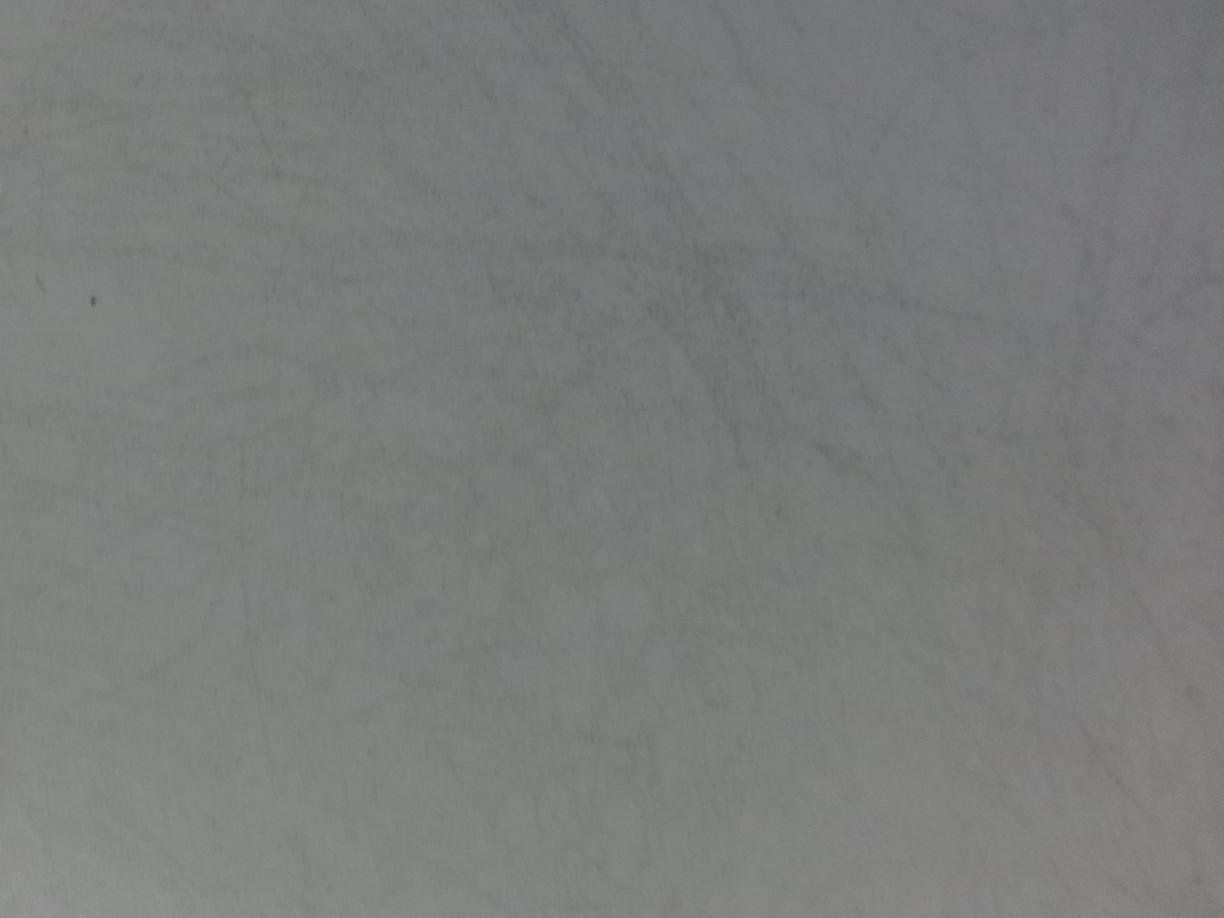}\end{overpic}
	    	\end{minipage}	
    }
	\label{subfig:samples-VisualNetWrong:18-19}
	\subfigure[Type 40 VS 52]{
			\begin{minipage}[b]{\colw\linewidth}
	    		\includegraphics[width=\figw\textwidth]{Experiment_figures/material/40/haptic_2.png}
	    		\begin{overpic}[width=\figw\textwidth]{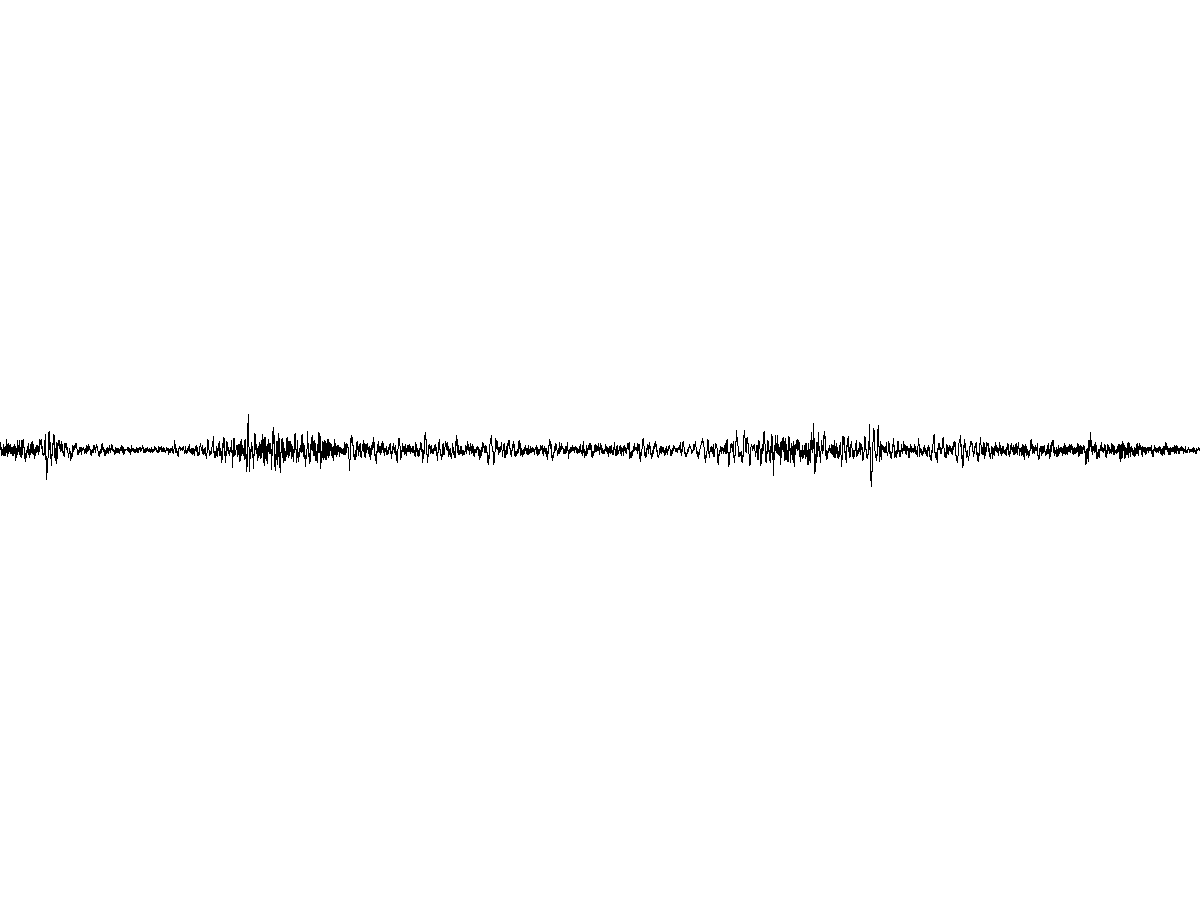}\end{overpic}\\
	    		\begin{overpic}[width=\figw\textwidth]{Experiment_figures/material/40/1_rezied.jpg}\end{overpic}
	    		\begin{overpic}[width=\figw\textwidth]{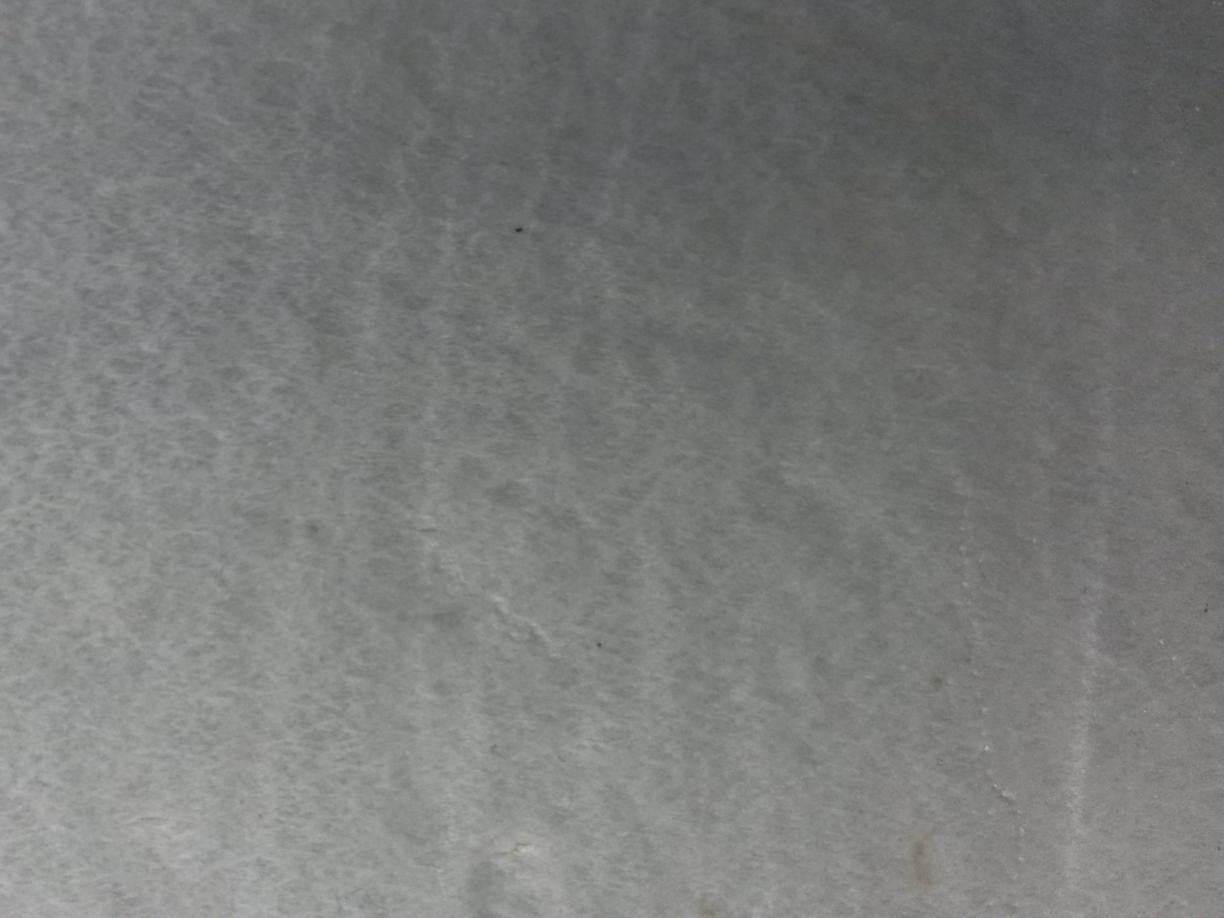}\end{overpic}
	    	\end{minipage}	
	}
	\label{subfig:samples-VisualNetWrong:40-52}
	\caption{\revised{(a)-(c) show the material types that \emph{HapticNet} cannot distinguish well, but \emph{VisualNet} can distinguish, where (a) is Type 12 (RoofTile) VS 13 (StoneTileVersion1), (b) is Type 39 (FineArtificialGrass) VS 40 (IsolatingFoilVersion1), (c) is Type 48 (FoamFoilVersion1) VS 64 (Leather). (d)-(f) depict the material types that \emph{VisualNet} cannot distinguish well, but \emph{HapticNet} can distinguish, where (d) is Type 13 (StoneTileVersion1) VS 14 (StoneTileVersion2), (e) is Type 18 (CeramicPlate) VS 19 (CeramicTile), (f) is Type 40 (IsolatingFoilVersion1) VS 52 (StyroporVersion1).}}
	\label{fig:samples-Wrong}
\end{figure*}

\begin{figure*}[htbp]
	\centering
	\subfigure[HapticNet]{\includegraphics[width = \colw\linewidth]{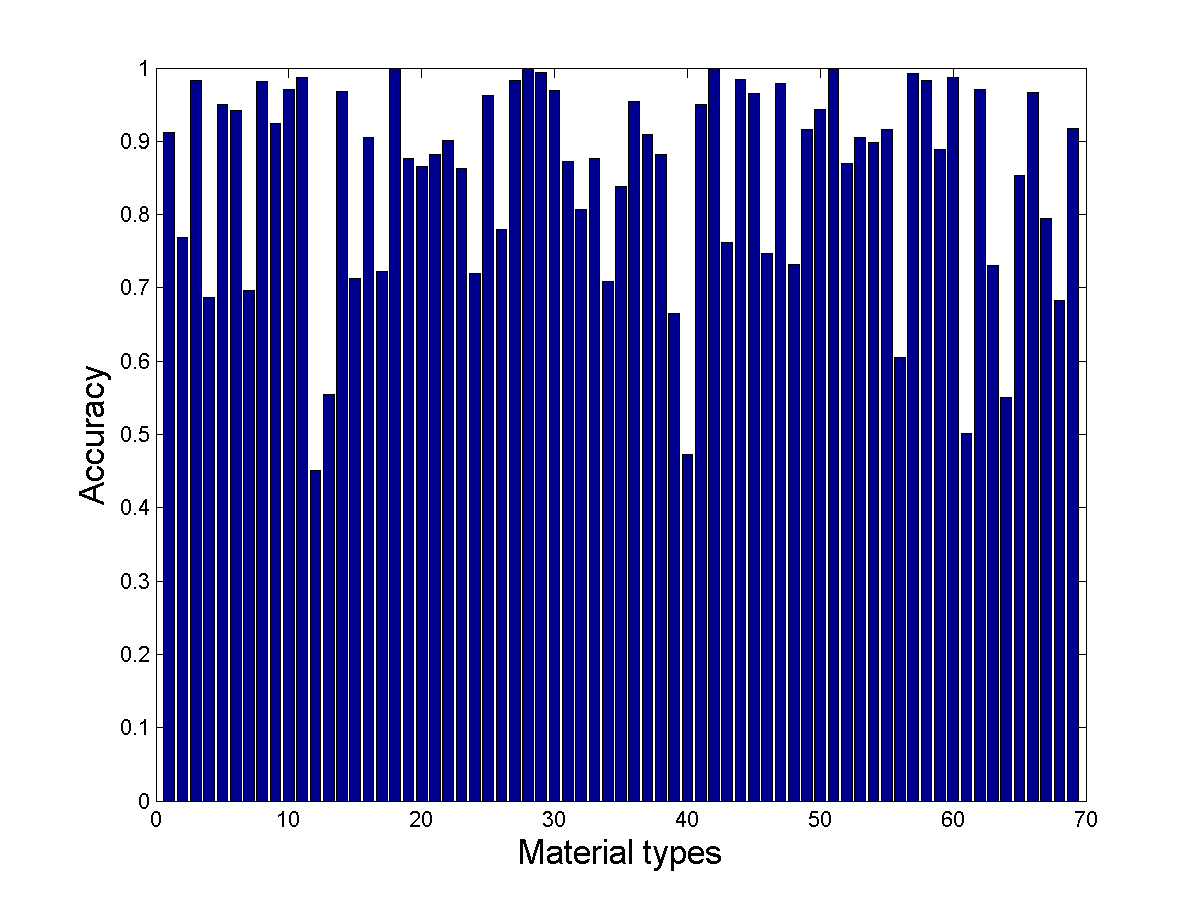}
	}
	\subfigure[VisualNet]{\includegraphics[width = \colw\linewidth]{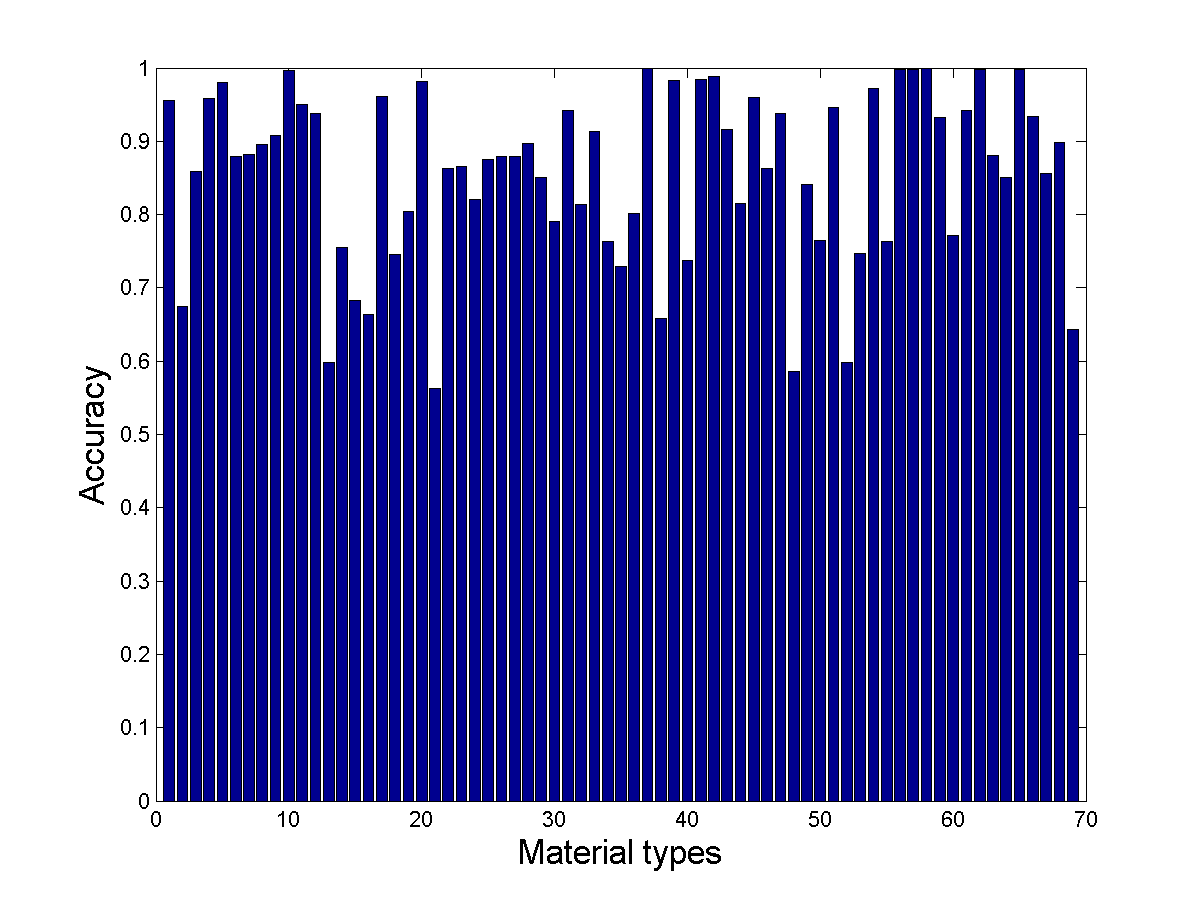}}
\subfigure[FusionNet-$\text{FC}_2$]{\includegraphics[width = 2in]{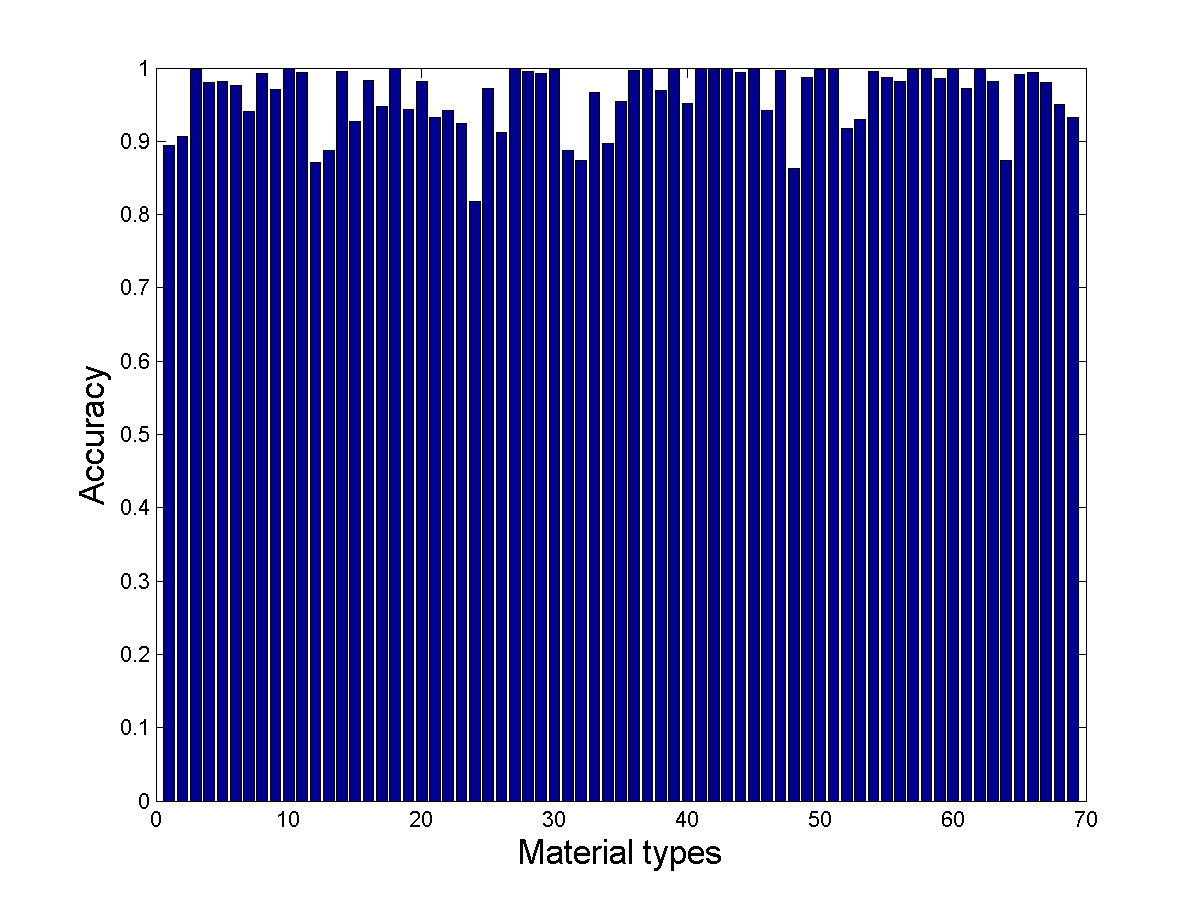}}
\caption{\revised{The fragment classification histogram (10-fold averages) of (a) \emph{HapticNet}; (b) \emph{VisualNet}; (c) \emph{FusionNet-$\text{FC}_2$}.}}	
\label{fig:histgram-fragment}
\end{figure*}

\vspace{0.3cm}
\noindent\textbf{Experiment Analysis}
\revised{To better understand the performance of our proposed scheme, we show the fragment classification confusion matrices of {\emph{HapticNet}}, {\emph{VisualNet}} and \emph{FusionNet-$\text{FC}_2$} in Fig. \ref{fig:confusion-matrices-track}, and the fragment classification accuracy is depicted in Fig. \ref{fig:histgram-fragment} as well.                           In the confusion matrices in Fig.~\ref{fig:confusion-matrices-track}, values between 0 to 1 are represented by colors varying from blue to red. Each row of the confusion matrices represents the probability of a material type being classified into the different 69 classes. Examining Fig. \ref{fig:confusion-matrices-track}(a) and Fig. \ref{fig:confusion-matrices-track}(b), we have following observations:
\begin{itemize}
  \item The off-diagonal misclassification patterns follow different distributions, implying that \emph{HapticNet} and \emph{VisualNet} have quite different behaviors when classifying surface materials.
  \item There are some material types which \emph{HapticNet} usually cannot distinguish well. For example: type 12 (RoofTile) and type 13 (StoneTileVersion1), type 39 (FineArtificialGrass) and type 40 (IsolatingFoilVersion1), type 48 (FoamFoilVersion1) and type 64 (Leather). However, they are more distinguishable using \emph{VisualNet}. Haptic/image samples of these materials are shown in Fig. \ref{fig:samples-Wrong}(a)-(c).
  \item Similarly, there are samples which \emph{VisualNet} cannot distinguish but \emph{HapticNet} can distinguish well. For example: type 13 (StoneTileVersion1) and type 14 (StoneTileVersion2), type 18 (CeramicPlate) and type 19 (CeramicTile), type 40 (IsolatingFoilVersion1) and type 52 (StyroporVersion1). Haptic/image samples of these materials are shown in Fig. \ref{fig:samples-Wrong}(d)-(f).
\end{itemize}}

\revised{Fig. \ref{fig:confusion-matrices-track}(c) shows that the above mentioned samples which were confused when working with a single modality, are mostly correctly classified for hybrid input. The diagonal line is more close to 1 compared to Fig. \ref{fig:confusion-matrices-track}(a) and (b). Different from Fig. \ref{fig:confusion-matrices-track}(a) and (b), fewer  off-diagonal values exist. \emph{FusionNet-$\text{FC}_2$} hence efficiently removes most misclassifications. The fragment classification accuracies of every material type in Fig. \ref{fig:histgram-fragment} also demonstrate the effectiveness of haptic/visual fusion, as revealed by the fact that \emph{FusionNet-$\text{FC}_2$} achieves stably high accuracy (vertical axis) when classifying all the 69 material types (horizontal axis).}

\vspace{0.3cm}
\noindent\textbf{Time profiling} To demonstrate the efficiency of our algorithm, we further compare the time profiling of \emph{HapticNet}/\emph{VisualNet} with two related `CNN + sliding window' approaches. For the  sliding window approaches, we transfer the weight of \emph{HapticNet}/\emph{VisualNet} to two Convolutional Neural Networks which share the same layer settings with \emph{HapticNet}/\emph{VisualNet}, while the $1 \times 1$ convolutional layer is replaced by a fully-connected layer. Clearly, with exactly the same weight, the sliding window approaches achieve the same fragment/max-voting accuracy as \emph{HapticNet} and \emph{VisualNet}. However, because FCN are able to avoid the redundant feature computation that exists in CNN + sliding window approaches, it should be faster. To validate that FCN based approaches like \emph{HapticNet}/\emph{VisualNet} are faster, we profile the average running time of \emph{HapticNet}/\emph{VisualNet} in comparison to CNN + sliding window approaches on a Nvidia GTX 860M graphics card. As shown in Table \ref{table:profiling}, for haptic classification, the sliding window-based approach takes 195.4ms to run, while \emph{HapticNet} only takes 20.5ms. Moreover, for visual classification, the sliding window approach takes 7903.4ms, while \emph{VisualNet} only takes 154.3ms.

\begin{table}[h!]
    \begin{center}
        \begin{tabular}{c|c}
        \hline\hline
        \emph{Haptic classification}                 & \emph{Average running time}  \\
        \hline
        Haptic CNN + sliding window            & 195.4 ms \\
        \textbf{{HapticNet}}                    & \textbf{20.5 ms}  \\

        \hline\hline

        \emph{Visual classification}                & \emph{Average running time}  \\
        \hline
        Visual CNN + sliding window            & 7903.4 ms  \\
        \textbf{{VisualNet}}                    & \textbf{154.3 ms}  \\
        \hline\hline
        \end{tabular}
        \caption {Profiling comparison between the proposed {\emph{HapticNet}} / \emph{VisualNet} and CNN + sliding windows approaches.}
        \label{table:profiling}
    \end{center}
\end{table}

\section{Conclusion and Future Work}
\label{sec:conclude}
We introduce a surface material classification method which uses Fully Convolutional Neural Networks. For predicting individual haptic or image input, we apply FCN with max-voting framework. We then design a fusion network dealing with both haptic and image input. Experiments on the TUM Haptic Texture Database demonstrate that our proposed system can achieve competitive classification accuracy compared to the existing schemes at reduced complexity. In our future work, we are aiming at further extending the current FCN + max-voting and hybrid classification schemes to more input types (images, haptic acceleration signals, sound signals, further signals from other modalities) for further improving the flexibility and robustness of our system.


\end{document}